\documentclass[]{imag-ms-template}
\usepackage{multirow}
\usepackage{hyperref}
\DeclareMathOperator*{\argmin}{arg\,min}

\usepackage{booktabs, makecell}
\usepackage{arydshln}
\usepackage{adjustbox}
\usepackage{xcolor}
\usepackage{lineno}
\title{Polaffini: A feature-based approach for robust affine and polyaffine image registration} 

\author{Antoine Legouhy$^{1,2,\ast}$, Cosimo Campo$^{1}$, Ross Callaghan$^{3}$,\\ Hojjat Azadbakht$^{3}$, Hui Zhang$^{1}$\\
{\small $^{1}$Hawkes Institute \& Department of Computer Science, University College London,
London, UK, }\\
{\small $^{2}$ Institut Pasteur, Université Paris Cité, Unité de Neuroanatomie Appliquée et Théorique, France, }\\
{\small $^{3}$AINOSTICS ltd., Manchester, UK}
}
\date{}
\begin{document} 

\maketitle 


\begin{abstract}

    In this work we present \emph{Polaffini}, a robust and versatile framework for anatomically grounded registration. 
    Medical image registration is dominated by intensity-based registration methods that rely on surrogate measures of alignment quality. In contrast, feature-based approaches that operate by identifying explicit anatomical correspondences, while more desirable in theory, have largely fallen out of favor due to the challenges of reliably extracting features. However, such challenges are now significantly overcome thanks to recent advances in deep learning, which provide pre-trained segmentation models capable of instantly delivering reliable, fine-grained anatomical delineations. We aim to demonstrate that these advances can be leveraged to create new anatomically-grounded image registration algorithms. To this end, we propose \emph{Polaffini}, which obtains, from these segmented regions, anatomically grounded feature points with 1-to-1 correspondence in a particularly simple way: extracting their centroids. These enable efficient global and local affine matching via closed-form solutions. Those are used to produce an overall transformation ranging from affine to polyaffine with tunable smoothness. Polyaffine transformations can have many more degrees of freedom than affine ones allowing for finer alignment, and their embedding in the log-Euclidean framework ensures diffeomorphic properties.
    \emph{Polaffini} has applications both for standalone registration and as pre-alignment for subsequent non-linear registration, and we evaluate it against popular intensity-based registration techniques. Results demonstrate that \emph{Polaffini} outperforms competing methods in terms of structural alignment and provides improved initialisation for downstream non-linear registration. \emph{Polaffini} is fast, robust, and accurate, making it particularly well-suited for integration into medical image processing pipelines.    
	Our code is freely available with documentation at~\url{https://github.com/CIG-UCL/polaffini}.
\end{abstract}

\section{Introduction}
    Medical image registration, the task of finding the best transformation to align one image with another such that their anatomical structures match, is an important tool in medical image analysis~\citep{hajnal2001,zhang2021,sotiras2013}. Subject-to-template registration, a.k.a. spatial normalisation, is a crucial step in most neuroimaging pipelines to enable population-level statistical analysis~\citep{gholipour2007,friston1995,brett2002}. Subject-to-subject registration is typically used in atlasing to produce unbiased population references~\citep{guimond2000,serag2012,joshi2004}.
    
    Methods for image registration can be broadly divided into feature-based and intensity-based strategies. In feature-based registration~\citep{malandain1993,thirion1996,declerck1995,pelizzari1989}, the correspondences (common features) between the two images are first extracted, then the transformation that best spatially aligns the corresponding features is estimated. These features are often based on geometric or anatomical characteristics, and the quality of alignment is assessed through spatial distances between corresponding features, which provide inherent interpretability. However, the practical utility of these approaches has historically been limited because extracting these features required tedious expert annotations or computationally intensive, yet inaccurate, automatic algorithms. In addition, portability is limited because the nature of the features and the extraction strategy must be carefully tailored to the modality and anatomy considered. As a result, the popularity of feature-based registration has declined.
    
    In contrast, intensity-based methods~\citep{shen2002,rueckert1999,thirion1998,Avants2008,Jenkinson2002,maes1997} circumvent the need to extract correspondences by relying on a similarity measure between voxel-wise intensities of the two images as a surrogate criterion of alignment quality. This reduces the registration problem to a simple and general formulation in which a similarity measure is maximised with respect to the transformation parameters, making the solution approachable by calculus techniques such as gradient descent. The similarity measure is chosen with respect to the relationship between the intensities of the images to be compared~\citep{roche1999}; an evaluation of similarity measures for registration tasks can be found in~\cite{avants2011}. These approaches now largely dominate the field~\citep{Oliveira2016,zhang2021}, several of which are evaluated in~\cite{klein2009}. However, these methods inherently lack interpretability because they do not explicitly establish anatomical correspondences.
    
    In addition, the similarity measure landscape with respect to the transformation parameters is highly non-convex, making the optimisation process susceptible to local minima. Avoiding these local minima requires a good initial estimate of the target transformation. For non-linear registration, a common strategy is to initialise the non-linear transformation using the result of a preceding linear registration, which is less prone to local minima due to its more constrained parameter space. However, linear registration itself is not entirely immune to local minima~\citep{jenkinson2001}, motivating the development of hybrid approaches, such as block matching~\citep{ourselin2001,commowick2012a,commowick2012b,modat2014} that extracts corresponding features through intensity-based local comparisons of image patches. Nevertheless, determining a reliable starting point remains challenging.
    
    Recent advances in deep learning have sparked renewed momentum across many fields of science, and medical image registration is no exception. The most significant impact has been the development of ultra-fast, intensity-based methods~\citep{Balakrishnan2019,devos2017}, which leverage deep learning to eliminate the iterative, and therefore time-consuming, optimisation of the similarity metric, replacing it with direct and nearly instantaneous prediction of the transformation parameters; reviews can be found in~\cite{Chen2021} and \cite{haskins2020}. However, these existing applications of deep learning do not address the issue of interpretability inherent to intensity-based methods.

    In this paper, we argue that deep learning can instead be exploited to revive feature-based registration, leveraging its capability to overcome the challenges traditionally associated with feature extraction. In particular, deep-learning architectures have demonstrated remarkable success in medical image segmentation, arguably the area where the technology has reached its highest level of maturity within medical image analysis. Convolutional neural networks, such as U-Net, now dominate segmentation benchmarks and challenges~\citep{Isensee2021}, most notably for brain segmentation tasks. For instance, FastSurfer~\citep{Henschel2020} can accurately replicate FreeSurfer's segmentation output~\citep{Fischl2002,Fischl2004}, which traditionally takes over five hours, in under a minute. SynthSeg~\citep{Billot2023} further extends this capability with a model that is both contrast- and resolution-agnostic, enabling robust segmentation across a wide range of imaging protocols. These models are robust, well-validated, and available pre-trained off-the-shelf. Anatomical feature extraction, historically difficult to achieve, is now accessible at minimal cost.

    As an illustrative example, we present \emph{Polaffini}, a registration framework that leverages these easily accessible segmentations in a particularly simple way: the centroids of the segmented regions are chosen to be the anatomically grounded feature points (or keypoints) with 1-to-1 correspondence.  The corresponding centroids enable nearly instantaneous, global and local, affine matching via closed-form solutions. The transformations from local affine matching are fused to form a dense, overall transformation through the log-Euclidean polyaffine transformation (LEPT) framework from~\cite{Arsigny2009}, which ensures desirable mathematical properties such as diffeomorphism and easy inversion. The overall transformation can be a simple global affine or a significantly more flexible polyaffine with tunable smoothness, whose degrees of freedom are related to the number of segmented regions. 
    % A Log-Euclidean Polyaffine Registration for Articulated Structures in Medical Images Miguel ´Angel Mart´ın-Fern´andez, Marcos Mart´ın-Fern´andez, and Carlos Alberola-L´opez
    
    This makes \emph{Polaffini} ideal for initialising non-linear registration and for tasks requiring spatial normalisation while preserving the relative sizes of anatomical structures (e.g., in neurodegeneration studies).
    The implementation of \emph{Polaffini} is provided open source and integrates seamlessly into modern medical image processing pipelines, much like how in recent versions of FreeSurfer's pipeline, intracranial volume estimation and skull stripping are powered by deep-learning segmentation models (SynthSeg~\citep{Billot2023} and SynthStrip~\citep{hoopes2022} respectively).
    
    A preliminary version of this work was published in~\cite{Legouhy2023}. Simultaneously yet independently, a similar approach was proposed in EasyReg~\citep{Iglesias2023} to estimate an affine transformation using centroids of segmented regions, but without strategies to estimate a more flexible polyaffine one. In~\cite{Legouhy2023}, we evaluated the proposed method primarily as a pre-alignment step to improve the performance of subsequent traditional and deep-learning-based non-linear registration. In this extended version, we present a substantially expanded experimental analysis. We compare \emph{Polaffini} with affine registration tools from widely used software packages, including ANTs, FSL, NiftyReg, and Anima. The evaluation includes anatomical structure overlap after registration, failure rates, and an analysis of the geometric differences between the estimated affine transformations. Additionally, we assess the alignment quality of the polyaffine variant of \emph{Polaffini} across a range of smoothness parameters.

\section{Methods}
\label{method}

In this section, we first introduce polyaffine transformations which can capture non-global deformations, enabling finer alignment, while remaining computationally efficient (Section~\ref{polyaffine}). We then detail the \emph{Polaffini} framework (Section~\ref{recipe}), a key component of which is a graph built from the centroids to account for neighbourhood information, allowing estimation of an affine transformation per neighbourhood in a way that preserves structural consistency between regions. At the end, we discuss theoretical and practical considerations of the proposed method (Section~\ref{considerations}).

\subsection{Polyaffine transformations} \label{polyaffine}

\begin{figure}[h!]
        \centering
        \includegraphics[width=0.9\linewidth]{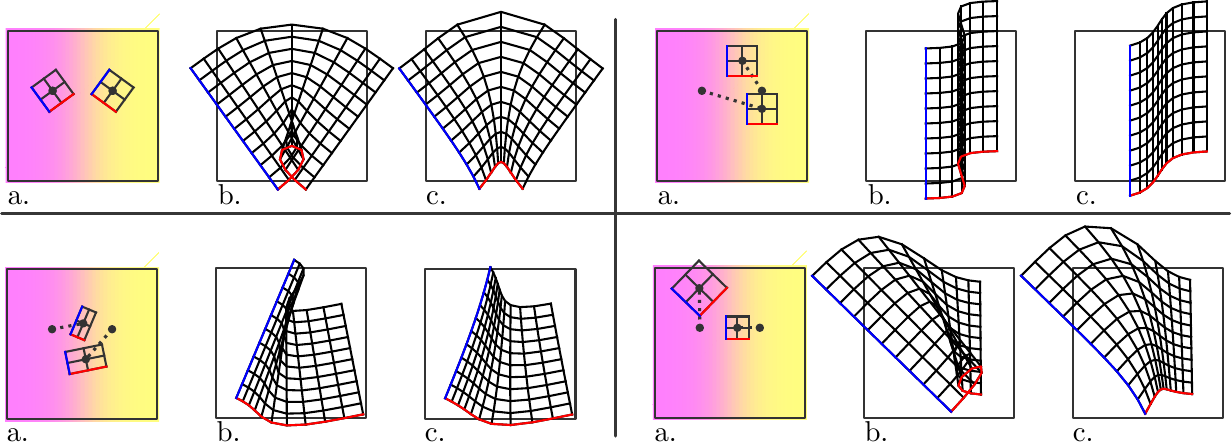}
        \caption{Fusion of multiple affine transformations. For each quadrant: a. local affine transformations and their associated weight maps; b. the resulting transformation obtained by naïve fusion through Euclidean average; and c. the resulting transformation obtained using the log-Euclidean polyaffine framework of~\cite{Arsigny2009} (illustration inspired by that paper).}
        \label{polyaff_eg}
    \end{figure}

    In this paper, the term polyaffine transformation specifically refers to the log-Euclidean polyaffine transformation (LEPT), as introduced in~\cite{Arsigny2009}. This framework enables the fusion of a set of local affine transformations, each attached to a feature point, into an overall transformation defined on the whole ambient space that retains desirable mathematical properties.
    The contribution of each local affine transformation to the final deformation is modulated by smooth weight maps, based on spatial proximity to the corresponding feature point.
    A naïve arithmetic average of affine matrices does not, in general, yield an invertible or geometrically meaningful transformation. Instead, the Lie group structure of the affine transformations is leveraged by computing the principal matrix logarithm of each one, mapping them into the Lie algebra, where linear combinations are well-defined and stable. These log-affine elements are then combined through a weighted arithmetic average, using spatial weight maps, to produce a stationary velocity field (SVF).
    The SVF is integrated, through the log-Euclidean framework for diffeomorphisms~\citep{Arsigny2006b}, resulting in a diffeomorphic transformation. This integration can be performed efficiently using the scaling and squaring method. The inverse of the polyaffine transformation can be generated simply by keeping the same weights and inverting each local affine transformation (or negating the SVF before integration). Examples of polyaffine transformations are shown in Fig.~\ref{polyaff_eg}.
    
\subsection{Polaffini framework} \label{recipe}

    The Polaffini framework we propose consists of seven steps. We take as input fine-grained segmentation of two images to be registered (step~\ref{seg}). We extract the centroids of the segmented regions to obtain anatomically grounded feature points (step~\ref{fpoints}). Using these paired sets of feature points, we estimate a background affine transformation (step~\ref{affonly}). If only an affine transformation is required, one can stop here; this part of the algorithm is identical to the method independently developed in~\cite{Iglesias2023}.
    
    To compute a polyaffine transformation, we need five additional steps. A set of corresponding local neighbourhood pairs is determined (step~\ref{graph}). For each pair of corresponding neighbourhoods, a local affine transformation is estimated (step~\ref{locaff}). Weight maps are established to spatially modulate the contribution of each local affine transformation (step~\ref{wmap}). A stationary velocity field (SVF) is estimated by weighted average of the logarithms of the local affine matrices (step~\ref{svf}). The SVF is integrated to produce a polyaffine transformation (step~\ref{intdiffeo}).  The details of each step are described below.
    
\begin{figure}[h!]
        \centering
        \includegraphics[width=\linewidth]{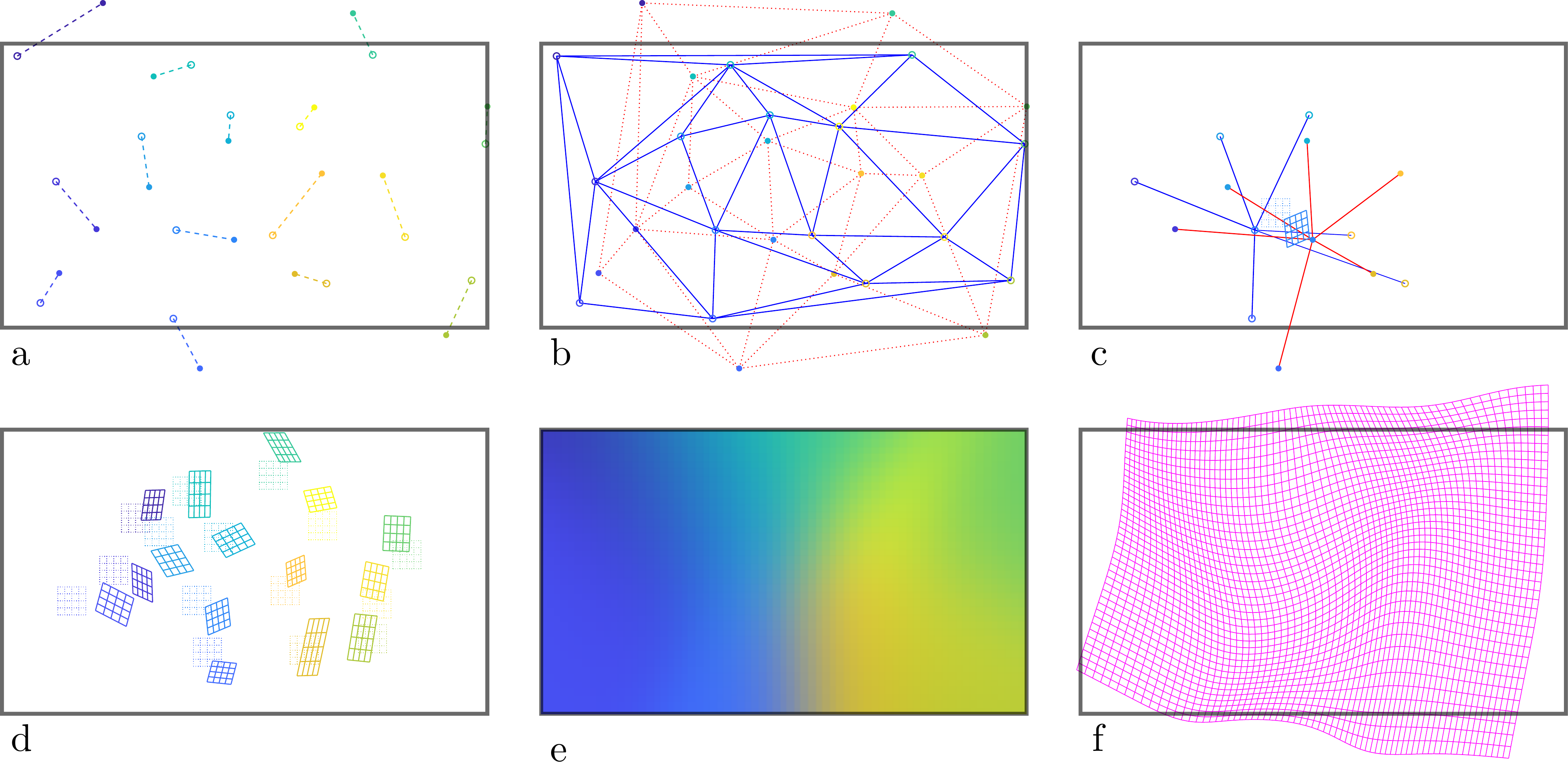}
        \caption{Illustration of several steps of the polyaffine estimation. a) Two sets of paired points (plain disc: moving, circle: reference). b) Delaunay tetrahedralisation performed on the reference set (blue) and pattern reproduced on the moving one (red). c) Local affine regression between two homologous neighbourhoods. d) Set of estimated local affine transformations. e) Colour gradient representing all weight maps combined. f) Overall polyaffine transformation. Black rectangles represent the frame of the reference image.}
        \label{diagfig}
    \end{figure}

    \begin{enumerate}
    \setcounter{enumi}{-1}
    \item\label{seg} \textbf{Segmentation:} A common fine-grained segmentation is performed on the reference and moving images using a pre-trained deep-learning segmentation model, such as FastSurfer~\citep{Henschel2020} or SynthSeg~\citep{Billot2023}. For details about the influence of the segmentation granularity and quality on the estimated transformation, see Section~\ref{influence_seg}.
    
    \item\label{fpoints} \textbf{Extraction of the feature points:}
    Given the fine-grained segmentations of the reference and moving images, the feature points (or keypoints) are defined as the centroids of the $n$ segmented regions. 
    This leads to two paired sets of points (see Fig.~\ref{diagfig}.a.):
    $$\left.\begin{array}{rl}
    \text{Reference point set:} & \{x_i\in \mathbb{R}^d\ |\ i=1,\dots,n\}\\
    \text{Moving point set:} & \{y_i\in \mathbb{R}^d\ |\ i=1,\dots,n\}
    \end{array}\right.\text{ ,\quad where } i \text{ indexes the regions.}$$
    
    After this point, the subsequent steps of the method are agnostic to how the feature points were obtained.
    
    \item\label{affonly} \textbf{Estimation of the background affine transformation:} The task of finding an optimal global affine transformation $\hat{A}_B$ mapping the two paired sets can be formulated in general as a weighted linear least squares (WLLS) regression problem: 
    \begin{equation}
    \label{llseq}
    \hat{A}_B=\left(\begin{array}{cc}\hat{L}_B & \hat{t}_B\\0& 1\end{array}\right)
    \text{\quad with \quad } (\hat{L}_B,\hat{t}_B)=\argmin_{\substack{L\in \mathbb{R}^{d\times d}\\ t\in \mathbb{R}^d}}\sum_{i=1}^n \alpha_i \|y_i-(Lx_i+t)\|_2^2 \end{equation}
    where $\alpha_i \geq 0$ denotes the known weight for region $i$, such that $\sum_{i=1}^{n} \alpha_i = 1$. This formulation is chosen because the resulting $\hat{A}_B$ maps reference coordinates to moving coordinates, which is the mapping required to resample the moving image onto the reference grid (backward mapping). There are a number of ways to choose the set of weights $\{\alpha_i\}$. In our preliminary work and in~\cite{Iglesias2023}, all regions are assigned equal weights of $1/n$. Alternatively, weights can be set proportional to each region’s relative size.
    
    This problem has a closed-form solution. If we define the mean locations of the reference and moving sets as $\bar{x}=\sum_{i=1}^n \alpha_i x_i$, $\bar{y}=\sum_{i=1}^n \alpha_i y_i$, the coordinates relative to the mean locations can be computed as $x'_i=x_i-\bar{x}$ and $y'_i=y_i-\bar{y}$. The closed-form solution can then be determined from these relative coordinates in a straightforward way:
    \begin{equation}
    \label{solllseq}
    \hat{L}_B=\sum_{i=1}^n \alpha_i y'_i{x'_i}^T\left(\sum_{i=1}^n \alpha_i x'_i{x'_i}^T\right)^{-1} \text{\quad and\quad  } \hat{t}_B=\bar{y}-\hat{L}_B\bar{x}
    \end{equation}
    
    The variant of \emph{Polaffini} that stops at this stage, as in~\cite{Iglesias2023}, is hereafter referred to as \emph{Polaffini-aff}.

    For subsequent local estimation, the moving points are pre-aligned using the inverse background affine transformation, yielding $\left\{\tilde{y_i}=\hat{A}_B^{-1}y_i = \hat{L}_B^{-1}(y_i -\hat{t}_B)\ |\ i=1,\dots,n\right\}$.
    
    \item\label{graph} \textbf{Determining the local neighbourhoods:} Estimating local transformations require pairs of corresponding local neighbourhoods. There are numerous ways to partition the space into such neighbourhoods, which may vary in number, size, and degree of overlap. At one extreme, each feature point may be treated as its own distinct neighbourhood. In this case, the correspondence between two neighbourhoods reduces to a correspondence between a single pair of feature points which is sufficient to estimate only a translation. Estimating a local affine transformation requires larger neighbourhoods that support multiple correspondences. Each neighbourhood must contain a sufficient number of affinely independent point pairs to constrain an affine model (at least four in 3D, three in 2D), while remaining sufficiently small to preserve locality.
        
    A convenient way to define such neighbourhoods is to build a graph structure that connects points based on spatial proximity. This graph-based grouping defines, for each point $x_i$, a neighbourhood $\mathcal{N}(x_i)$ consisting of the point itself and the points directly connected to it (see Fig.~\ref{diagfig}.b and ~\ref{diagfig}.c). The graph should be chosen such that each neighbourhood contains an adequate number of points for affine estimation. Delaunay tetrahedralisation in 3D (triangulation in 2D) provides an efficient way to satisfy those requirements.
  
    Because the moving and reference point sets are in one-to-one correspondence, the neighbourhoods only need to be determined for one point set, then transferred via the correspondence to the other: $x_p\in \mathcal{N}(x_i)\iff \tilde{y}_p\in \mathcal{N}(\tilde{y}_i)$. We therefore denote neighbourhoods by index sets $\mathcal{N}(i)$, where $p\in \mathcal{N}(i)$ indicates that point $p$ is connected to point $i$ in the graph. In our implementation, we choose to determine the neighbourhoods with the reference point set, because this choice is particularly convenient for spatial normalisation. In this setting, the neighbourhoods (for the template) only needs to be computed once for the whole population.
 
    \item\label{locaff} \textbf{Estimation of the local affine transformations:} For each neighbourhood index $i$, a local affine transformation \(\hat{A}_i\) best aligning the reference neighbourhood \(\{x_p \mid p\in\mathcal{N}(i)\}\) to its corresponding (pre-aligned) moving one \(\{\tilde{y}_p \mid p\in\mathcal{N}(i)\}\) is sought (see Fig.~\ref{diagfig}.c). The situation is similar to step~\ref{affonly} except that only the points of the neighbourhood are used. Similarly to~Eq.~\ref{llseq}, the local WLLS problem is: 
    \begin{equation}
    \label{affreg}
    \hat{A}_i=\left(\begin{array}{cc}\hat{L}_i & \hat{t}_i\\0& 1\end{array}\right)
    \text{\quad with \quad } (\hat{L}_i,\hat{t}_i)=\argmin_{\substack{L\in \mathbb{R}^{d\times d}\\ t\in \mathbb{R}^d}} \sum_{p\in \mathcal{N}(i)} \alpha_p\|\tilde{y}_p-(Lx_p+t)\|^2_2
    \end{equation}

    Let us consider $\bar{x}_i=\frac{1}{\sum_{p\in \mathcal{N}(i)}\alpha_p}\sum_{p\in \mathcal{N}(i)}\alpha_px_p$, $\bar{\tilde{y}}_i=\frac{1}{\sum_{p\in \mathcal{N}(i)}\alpha_p}\sum_{p\in \mathcal{N}(i)}\alpha_p\tilde{y}_p$, and the relative coordinates $x'_p=x_p-\bar{x}_i$ and $\tilde{y}'_p=\tilde{y}_p-\bar{\tilde{y}}_i$. Similarly to Eq.~\ref{solllseq}, the solution is: 
    \begin{equation}
    \label{solllsloc}
    \hat{L}_i=\sum_{p\in \mathcal{N}(i)}\alpha_p\tilde{y}'_p{x'_p}{}^T\left(\sum_{p\in \mathcal{N}(i)}\alpha_px'_p{x'_p}{}^T\right)^{-1} \text{\quad and\quad  } \hat{t}_i=\bar{\tilde{y}}_i-\hat{L}_i\bar{x}_i
    \end{equation}
    
    At this stage we have $n$ local affine transformations between corresponding neighbourhoods attached to the $n$ reference feature points (see Fig.~\ref{diagfig}.d).
    
    \item\label{wmap} \textbf{Creation of the weight maps:}
    To create an overall dense transformation, smooth weight maps are established to spatially modulate the contribution of each local affine transformation. For any point $x$ of the domain over which the image is defined, a set of weight maps $\{w_i(x),\enskip i=1,\dots,n\}$ associated with each $x_i$ is defined using a smooth kernel function based on the distance between $x$ and the center of the associated neighbourhood $\bar{x}_i$, and a parameter $\sigma$ controlling the smoothness (see Fig.~\ref{diagfig}.e). The weight maps can typically be produced using Gaussian kernels of standard deviation $\sigma$:
    \begin{equation}
    w_i(x)=\exp\left(-\frac{1}{2}\frac{\|x-\bar{x}_i\|^2_2}{\sigma^2}\right)
    \end{equation}
    To improve stability, a background weight $w_B$, uniform across the whole image domain, is also chosen: $w_B(x)=w_B$. It should be small enough to be negligible with respect to the sum of the other weights when getting close to the feature points (see Section~\ref{rmkwb} for more details).
    
    \item\label{svf} \textbf{Estimation of a stationary velocity field:} From the collection of local affine transformations and the weight maps, one can produce an overall dense diffeomorphic transformation through the log-Euclidean polyaffine framework from~\cite{Arsigny2009}. 
    A stationary velocity field (SVF) $V$ is built by interpolating a log displacement vector for each $x$ by averaging the logarithms of the local transformations, weighted by the associated weight maps and the background weight:
    \begin{equation}\label{logmean}
    V(x)=\frac{\sum_{i=1}^n w_i(x)\cdot\log(\hat{A}_i)}{w_B+\sum_{i=1}^nw_i(x)}x
    \end{equation}
    Here, $\log(\hat{A}_i)$ denotes the matrix logarithm of the affine transformation $\hat{A}_i$, not the element-wise logarithm. See Section~\ref{logmat} for more detail about the computation, existence and uniqueness of those logarithms.
    
    \item\label{intdiffeo} \textbf{Integration onto a diffeomorphic transformation:}

    The log-Euclidean framework for~\citep{Arsigny2006b} diffeomorphisms confers an infinite Lie group structure to diffeomorphisms parametrised by SVF. A sufficiently smooth SVF $V$ defines a one-parameter subgroup of diffeomorphisms $\varphi^t$, given as the unique solution of the stationary ordinary differential equation:
    \begin{equation}
    \label{ode}
     \dfrac{\partial \varphi^t}{\partial t}=V(\varphi^t)
     \text{\quad with initial condition\quad } \varphi^0=\mathrm{Id}
    \end{equation}

    Integrating this equation during one unit of time defines the Lie exponential map, which yields the final transformation (see Fig.~\ref{diagfig}.f): 
    \begin{equation}
    \exp(V)=\varphi^1 \text{\quad simply noted\quad }  \varphi
    \end{equation}
    And the inverse of the transformation can simply be computed through: $\varphi^{-1}=\exp(-V)$.
    The integration can be done efficiently on regular grids using the scaling and squaring method~\citep{Arsigny2006b,Arsigny2009} by approximating the exponential of the scaled (close to 0) field and composing recursively. 

    The final transformation $T$, mapping the moving image to the reference one, can then be obtained by composing the background affine and polyaffine transformations:
    \begin{equation}T=\hat{A}_B\circ \varphi\end{equation}
        
    \end{enumerate}

    \subsection{Theoretical and practical considerations}\label{considerations}

    This section outlines additional theoretical underpinnings of \emph{Polaffini}, along with recommendations for implementation and guidance on its integration into medical image processing pipelines.
  
    % \subsubsection{}
    % For more robustness, one can imagine weighted or trimmed versions of the linear least squares minimisation in Eq.~\ref{affreg} if the graph defined at step~2 is weighted, or to account for uncertainty in the feature point extraction.

    \subsubsection{Influence of the quality of the segmentation}
        \label{influence_seg}
        Since only the centroids of the labelled regions are used, the boundary precision of the segmentation matters little, making \emph{Polaffini} remarkably robust to coarse, imperfect segmentations. However, the number of regions directly influences the flexibility potential of the polyaffine transformation. The effective degrees of freedom are roughly proportional to the number of labels, modulated by the smoothness parameter $\sigma$. With more regions, one can use smaller values of $\sigma$ and achieve finer alignment, or keep the same smoothness and have more constraints for improved robustness. The method is scalable to whatever granularity of segmentation is available, with the minimum requirement of 4 labels in common between the two images in 3D.
        
    \subsubsection{Local transformations type}
        In a $d$-dimensional space, at least $d+1$ points that are affinely independent (non-collinear in 2D, non-coplanar in 3D) are needed in each of the reference and moving neighbourhoods to fully constrain Eq.~\ref{llseq} and avoid degeneracy.

        Should the neighbourhoods be smaller, local transformations with fewer degrees of freedom can still be used. For a rigid transformation, i.e. $L$ constrained to be a rotation matrix, a minimum of $d$ affinely independent points in each neighbourhood is needed, and a direct solution to the LLS problem can be found in~\cite{Horn1987,Horn1988}. For translations only, i.e. $L$ is the identity matrix, only singleton neighbourhoods are necessary ($\mathcal{N}(x_i)=\{x_i\}$), and the solution reduces to $\hat{t}=\bar{y}-\bar{x}$. However, this setting can be prone to overfitting and the contextual information from the neighboring positions is discarded (although it still has an impact during interpolation).
 
    \subsubsection{Alternative affine transformation estimation}
        To ensure that the estimated affine matrix is invertible, one might consider instead of the WLLS problem in Eqs.~\ref{llseq}~and~\ref{affreg}, the following non-linear least squares alternative taking advantage of the Lie group structure of the set of invertible matrices:
        $$\hat{A}=\left(\begin{array}{cc}\exp(\hat{\Lambda}) & \hat{t}\\0& 1\end{array}\right)
        \text{\quad with \quad } (\hat{\Lambda},\hat{t})=\argmin_{\substack{\Lambda\in\mathbb{R}^{d\times d}\\ t\in \mathbb{R}^d}} \sum_{i=1}^n\alpha_i\|y_i-(\exp(\Lambda)x_i+t)\|^2$$
        This formulation leads to invertible affine transformation estimates even when dealing with degenerate or ill-conditioned data: $\exp(\Lambda)\in \operatorname{GL}(d,\mathbb{R})$ and $A\in \operatorname{GL}(d+1,\mathbb{R})$. However, unlike its WLLS counterpart, this problem does not have a known closed-form solution.
    
    \subsubsection{Smoothness of the polyaffine transformation} \label{rmkweightmap}
        The choice of the weight maps influences the shape of the overall transformation. The width of the kernel modulates the amount of smoothness. For small values of $\sigma$, one can capture local spatial variations but there is a risk of overfitting or having sub-machine precision values when the feature point set is too sparse in space (although the use of a background weight helps). As $\sigma$ increases, the polyaffine transformation gets smoother, at the cost of reduced locality. This flexibility allows the transformation to be adapted to the application and to the density of available feature points. 
        In the limiting case $\sigma=\infty$, all weight maps are uniform over the image domain and the resulting transformation reduces to an affine, log-Euclidean average of the local ones: $\hat{A}=\hat{A}_B\exp\left( \frac{1}{n+w_B}\sum_{i=1}^n\log(\hat{A}_i) \right)$. For $\sigma=0$ and $w_B>0$, the polyaffine transformation reduces to the affine background transformation only.
    
    \subsubsection{Role of the background transformation and weight} \label{rmkwb}
\begin{figure}
            \centering
            \includegraphics[width=\linewidth]{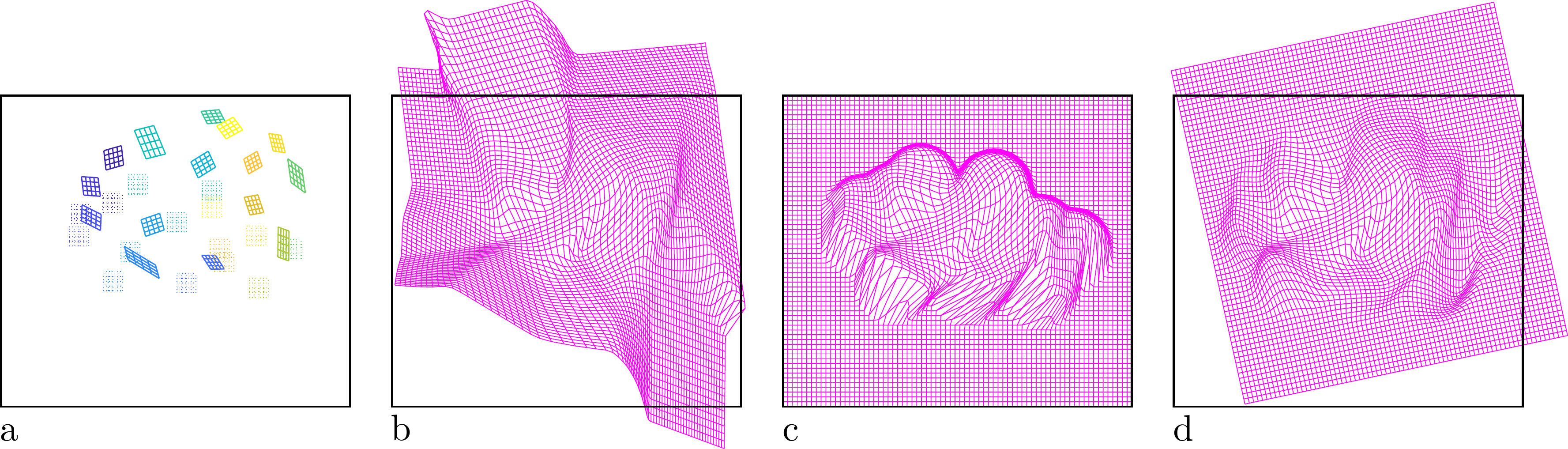}
            \caption{For a given set of local affine transformations (a), illustration of the influence of the choice of the background transformation for the construction of the overall polyaffine transformation:  b) no background transformation, c) identity background transformation. d) affine background transformation. Black rectangles represent the frame of the reference image.}
            \label{bgt}
        \end{figure}

        In estimating the overall polyaffine transformation, aberrant behaviours may occur when extrapolating far beyond the feature points (see Figure~\ref{bgt}.b). In particular, it is likely to encounter situations where the weights fall below numerical precision, resulting in indeterminacy when normalising them. This is especially true if the smoothness parameter $\sigma$ is small or if the image boundaries loosely encompass the set of points, leaving substantial margins.
        The role of the background transformation is to ensure stability in those (relatively to $\sigma$) remote areas. The influence of the background transformation is modulated by a background weight $w_B$, uniform across the whole image domain, that should be chosen sufficiently small to only matter far enough from the feature points but not interfere near them. As \emph{Polaffini} does not assume any pre-alignment, using an identity background transformation would not be appropriate (see Figure~\ref{bgt}.c). Instead, it should follow the overall flow in the smoothest way i.e. through a global affine background transformation (see Figure~\ref{bgt}.d). 

    \subsubsection{Principal logarithm of matrices}\label{logmat}
        In Eq.~\ref{logmean}, we compute an arithmetic mean of the principal logarithms of the local affine transformations.\\
        The existence and uniqueness of the principal logarithm of an affine transformation matrix $A$ only depends on its linear part $L$~\citep{Arsigny2009}. The eigenvalues of $L$ must not lie on the (closed) half-line of negative real numbers~\citep{hun2001,gallier2008}. 
        Violations of this condition correspond to transformations involving rotations near $\pi$, which are impractical cases in image registration tasks.\\
        Computationally, only $n$ matrices (one per local transformation) must be evaluated. These can be computed efficiently using the inverse scaling and squaring method, as detailed in~\citep{higham2008}. 

    \subsubsection{Lie group embedding}
        \emph{Polaffini} is embedded in the log-Euclidean framework for diffeomorphisms~\citep{Arsigny2006b}, which endows SVF-parametrised transformations with an infinite-dimensional Lie group structure. 
        This allows linear operations to be performed in the Lie algebra, with the result mapped back to a diffeomorphism via the Lie exponential. Since many non-linear registration algorithms also use SVF parametrisation, \emph{Polaffini} integrates seamlessly with them. 
        In particular, the Baker–Campbell–Hausdorff (BCH) formula can be used to directly approximate the SVF of a composed transformation from the SVFs of its components~\citep{Bossa2007, Vercauteren2009a}, avoiding the need for potentially expensive logarithm computations. This is especially useful in atlasing~\citep{schuh2014,legouhy2019}.

    \subsubsection{Polaffini as pre-alignment for non-linear registration}
        \begin{figure}[h!]
            \centering
            \includegraphics[width=\linewidth]{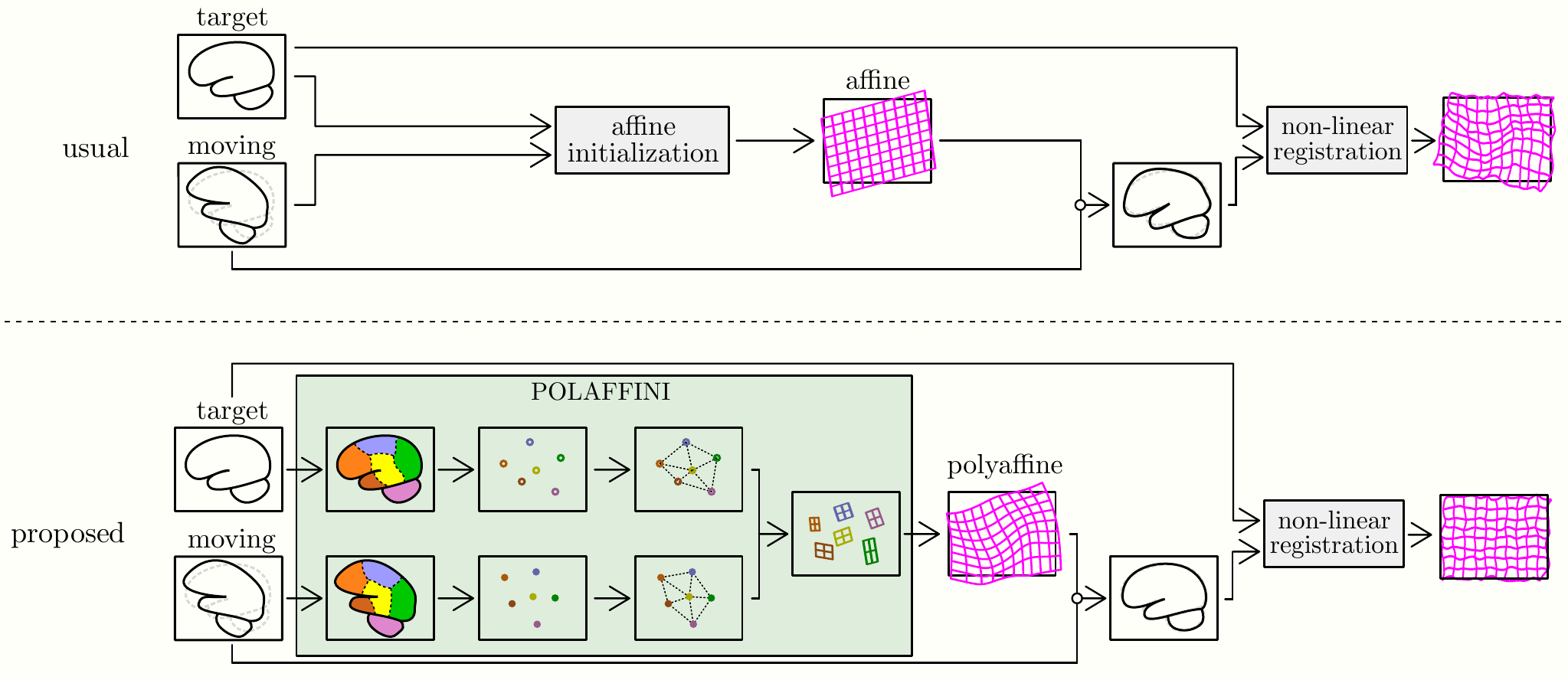}
            \caption{Usual (top) and proposed (bottom) paths for non-linear registration.}
            \label{pplfig}
        \end{figure}
        Most non-linear registration pipelines for mapping a moving image onto a target one (a template or another subject) follow three main steps (see Fig.~\ref{pplfig}, top part): firstly, an initial intensity-based affine registration is performed. Secondly, a non-linear registration is performed, using the output of the affine one as starting point. Finally, the two transformations are composed to produce the final transformation.
        Although the number of affine parameters to estimate is relatively small (12 in 3D), achieving robust performances is challenging. For the non-linear registration, the optimisation is difficult due to the large number of dofs and the highly non-convex optimisation landscape. The affine initialisation helps, but it is limited: it provides only global, inflexible transformations and relies on a surrogate similarity measure that does not directly reflect anatomical alignment.
        
        We propose to replace the affine intensity-based initialisation with \emph{Polaffini} (see Fig.~\ref{pplfig}, bottom part). \emph{Polaffini} estimates anatomically grounded transformations and can leverage a polyaffine model with significantly more degrees of freedom, enabling local deformations that allow finer and more accurate alignment. This matching is also very robust. \emph{Polaffini} should therefore provide a better starting point compared to its affine counterpart, thus offering better conditions for the convergence of the optimisation process.
        
        \emph{Polaffini} is especially well suited as a pre-alignment step for deep-learning registration models, such as VoxelMorph~\citep{Balakrishnan2019,devos2017}. Spatial normalisation reduces variability in the input domain, allowing the model to focus on relevant features rather than global alignment. This improves training efficiency, reduces overfitting, and enhances generalisation. For optimal performance, \emph{Polaffini} should be applied both at training for the above-mentioned learning benefits; and testing to avoid domain shift caused by inconsistent spatial distributions.

\section{Evaluation}

    We conducted a series of experiments to evaluate the performance of the proposed method. We studied the influence of the smoothness parameter for the polyaffine mapping (Section~\ref{sigmacomp}). We compared \emph{Polaffini} against popular intensity-based affine registration methods in terms of robustness (Section~\ref{robust}), anatomical structure alignment after registration (Section~\ref{regresults}). We studied the geometric differences between those competing methods and \emph{Polaffini-aff} (Section~\ref{geocomp}). Finally, we evaluated the ability of \emph{Polaffini} to improve downstream non-linear registration in both traditional and deep-learning settings (Section~\ref{nonlininit}).
    
    \subsection{Experimental setup}

    \subsubsection{Data}
    \label{data}
    We used T1-weighted images from 3 databases in order to cover various acquisition protocols and to have brains of different maturation and health conditions:
    \begin{itemize}
    \item ADNI\footnote{ADNI:~\url{adni.loni.usc.edu}}, the Alzheimer's Disease Neuroimaging Initiative~\citep{Petersen2010}, is a cohort of elderly subjects divided into cognitively normal (HC), with mild cognitive impairment (MCI), and with Alzheimer's disease (AD). 
    \item IXI dataset\footnote{IXI dataset:~\url{brain-development.org/ixi-dataset}} is composed of adult healthy subjects aged 20 years old or more.
    \item UK Biobank\footnote{UK Biobank:~\url{ukbiobank.ac.uk}} is a huge database of subjects from the UK, between 40 and 69 years old.
    \end{itemize}
    More information about these databases can be found in the Acknowledgements section.
    
    For all databases, the voxel size is around 1 mm isotropic. 
    Subjects have been drawn randomly from those databases and distributed into training, validation and testing sets following  Table~\ref{datasplit}. The testing set is the one actually used for evaluation in all experiments. The training and validation sets are only used for the training of the deep-learning non-linear registration model used in~Section~\ref{nonlininit}. 
    For subject-to-template registration, the reference for registration is an MNI template (2009c symmetrical) with 1 mm isotropic voxel size.
    For subject-to-template registration, each subject in the test set was used as the moving image once, while the reference image was randomly selected from a permutation of the same set, leading to 350 unique pairs to be registered.
    
    \begin{table}[]
    \centering
    \caption{Distribution of the subjects for the training, validation and testing sets.}
    \begin{tabular}{c||c|c|c}
    Database & Training set & Validation set & Testing set\\\hline\hline
    IXI   & 20  & 5  & 100 \\
    UK Biobank & 20   & 5  & 100   \\
    \begin{tabular}{c}ADNI\vspace{-0.1cm}\\ \scriptsize{(HC/MCI/AD)}\end{tabular} & \begin{tabular}{c}60\vspace{-0.1cm}\\ \scriptsize{(20/20/20)}\end{tabular}  & \begin{tabular}{c}15\vspace{-0.1cm}\\\scriptsize{(5/5/5)}\end{tabular} & \begin{tabular}{c}150\vspace{-0.1cm}\\\scriptsize{(50/50/50)}\end{tabular}
    \end{tabular}
    \label{datasplit}
    \end{table}
    
    \subsubsection{Segmentation}
    We used SynthSeg~\citep{Billot2023} (with the -\text{-}parc option) to quickly (in less than a minute) produce FreeSurfer-like segmentations of the moving and reference images into 98 anatomical regions following the Desikan–Killiany–Tourville (DKT) protocol depicted in~\citep{Klein2012,Desikan2006}. For evaluation purposes, we grouped the labels into three anatomical groups: sub-cortical (labels: 10, 11, 12, 13, 17, 18, 26, 28, 49, 50, 51, 52, 53, 54, 58 and 60), cortical (labels greater than 1000) and white matter (super labels 2 and 41). When computing metrics for these groups, we report the average over the individual labels within each group, without merging the regions into a single aggregated label.

    \subsubsection{Competing registration methods}
    \label{competmethods}
        To benchmark our method, we compared it against four widely used intensity-based affine registration tools, each employing different optimisation strategies (image-wise or block-matching) and similarity measures:
        \begin{itemize}
            \item \textbf{Flirt}~\citep{Jenkinson2002}, from the FSL\footnote{FSL: \url{https://fsl.fmrib.ox.ac.uk/}, RRID:SCR\_002823, version 6.0.5.2.} suite: It computes a similarity at the whole image level. The similarity metric used was the correlation ratio \citep{roche1998}, which assumes a functional relationship between the intensities of the registered images.
            % $$\frac{1}{\text{var}(J)}\sum_k\frac{n_k}{N} \text{var}(J_k)$$
            \item \textbf{ANTs-aff}~\citep{avants2011}, from the Advanced Normalisation Tools\footnote{ANTs: \url{http://stnava.github.io/ANTs/}, RRID:SCR\_004757, version 2.5.4.} (ANTs) suite: It computes a similarity at the whole image level. The similarity metric used was the mutual information~\citep{wells1996,maes1997}, which assumes a statistical relationship between the intensities of the registered images.
            \item \textbf{Anima-aff}~\citep{commowick2012a}, from the Anima\footnote{Anima: \url{https://anima.irisa.fr}, RRID:SCR\_017017, version 4.2.} suite: It follows a block matching strategy. The similarity metric is the squared Pearson correlation coefficient, which assumes positive or negative correlation between the intensities of homologous patches between the images.
            \item \textbf{Aladin}~\citep{modat2014}, from the NiftyReg\footnote{NiftyReg: \url{https://github.com/KCL-BMEIS/niftyreg}, RRID:SCR\_006593, version 1.5.77} suite: It follows a block matching strategy. The similarity metric is the Pearson correlation coefficient, which assumes a positive correlation between the intensities of homologous patches between the images.
            % $$\frac{1}{N}\sum_{x\in b_r}\frac{(b_r(x)-\mu_{b_{r}})(b_f(x)-\mu_{b_{f}})}{\sigma_{b_r}.\sigma_{b_f}}$$
        \end{itemize}
        For Flirt, the search range for angles was restricted to $[-30, 30]$ degrees for improved robustness. Otherwise, apart from specifying the transformation model, all tools were used with default parameters. Each affine registration was preceded by a rigid one with the same tool.
        
    \subsubsection{Polaffini implementations}

        We consider two versions of \emph{Polaffini}:
        \begin{itemize}
            \item \textbf{Polaffini-aff}: Only an affine transformation is estimated, which is the background affine transformation. It corresponds to stopping the recipe in Section~\ref{recipe} after step~\ref{affonly}. This is similar to what is done in~\cite{Iglesias2023}.
            \item \textbf{Polaffini-polyaff}: A polyaffine transformation is estimated following the full recipe in~Section~\ref{recipe}.
        \end{itemize}
        
        We opted for the following implementation details:
        \begin{enumerate}
        \item \textbf{Extraction of the feature points:} The feature points were extracted by computing the centroid of each segmented region, except for: the left and right cerebral white matter (labels 2 and 41) that are not specific enough, and the outer cerebrospinal fluid (label 24). 
        \item \textbf{Estimation of the background global transformation:} We estimated a background affine transformation using all feature points following Eq.~\ref{solllseq}. This is the end-point for \emph{Polaffini-aff}.
        \item \textbf{Construction of a graph structure:} The graph structure was constructed through Delaunay tetrahedralisation using the Qhull library.
        \item \textbf{Estimation of the local affine transformations:}  We estimated a local affine transformation for each neighbourhood using the neighbourhood points following Eq.~\ref{solllsloc}.
        \item \textbf{Creation of the weight maps:} When not stated otherwise, a Gaussian kernel of standard deviation $\sigma=15$ mm was used for the weight maps. This choice was motivated by the results of the experiment in Section~\ref{sigmacomp}. A background weight of $w_B=10^{-5}$, uniform across the whole image domain was chosen.
        \item \textbf{Estimation of a stationary velocity field:} We computed the SVF following Eq.~\ref{logmean} on the image grid downsampled by a factor 4 to quicken the subsequent exponentiation.
        \item \textbf{Integration onto a diffeomorphic transformation:} 
        The exponential of the SVF in Eq.~\ref{ode} was computed through scaling and squaring using 7 integration steps and resampled onto the original grid.
        \end{enumerate}
        
        We have made freely available on Github\footnote{\url{https://github.com/CIG-UCL/polaffini}} the implementation described here, which is based on the Python version of the SimpleITK wrapper for the ITK open-source software. 

    \subsubsection{Downstream non-linear registration methods}
        \label{nlintools}
        \begin{itemize}
            \item \textbf{Traditional non-linear registration:}
            Symmetric Normalisation (SyN)~\citep{Avants2008}, from the Advanced Normalization Tools (ANTs) suite was used. It is one of the best traditional non-linear algorithms according to the evaluation in~\cite{klein2009}. The optimised similarity metric was the local (over a sliding window) squared Pearson correlation coefficient (LCC).
        
            \item \textbf{Deep-learning non-linear registration:} A Voxelmorph-style architecture~\citep{devos2017,Balakrishnan2019} with diffeomorphic implementation~\citep{Dalca2019} and weak supervision with segmentations as auxiliary data~\citep{hu2018} was used.
            The U-Net and the resampler~\citep{jaderberg2015} were implemented like in~\cite{Balakrishnan2019}; the integration block similar to~\citep{Dalca2019}. We chose LCC as image similarity loss with weight 1, average Dice as segmentation loss with weight 0.3 and the L$^2$ norm of the Jacobian of the pre-integrated field as regularisation loss with weight 1. Due to limited GPU memory, the models operated on images downsampled to $2\times 2\times 2$ mm grids. 
        \end{itemize}
        
    \subsection{Experiments and results}
    
    \subsubsection{Influence of the smoothness parameter} \label{sigmacomp}
\begin{figure}
            \centering
            \includegraphics[width=\linewidth]{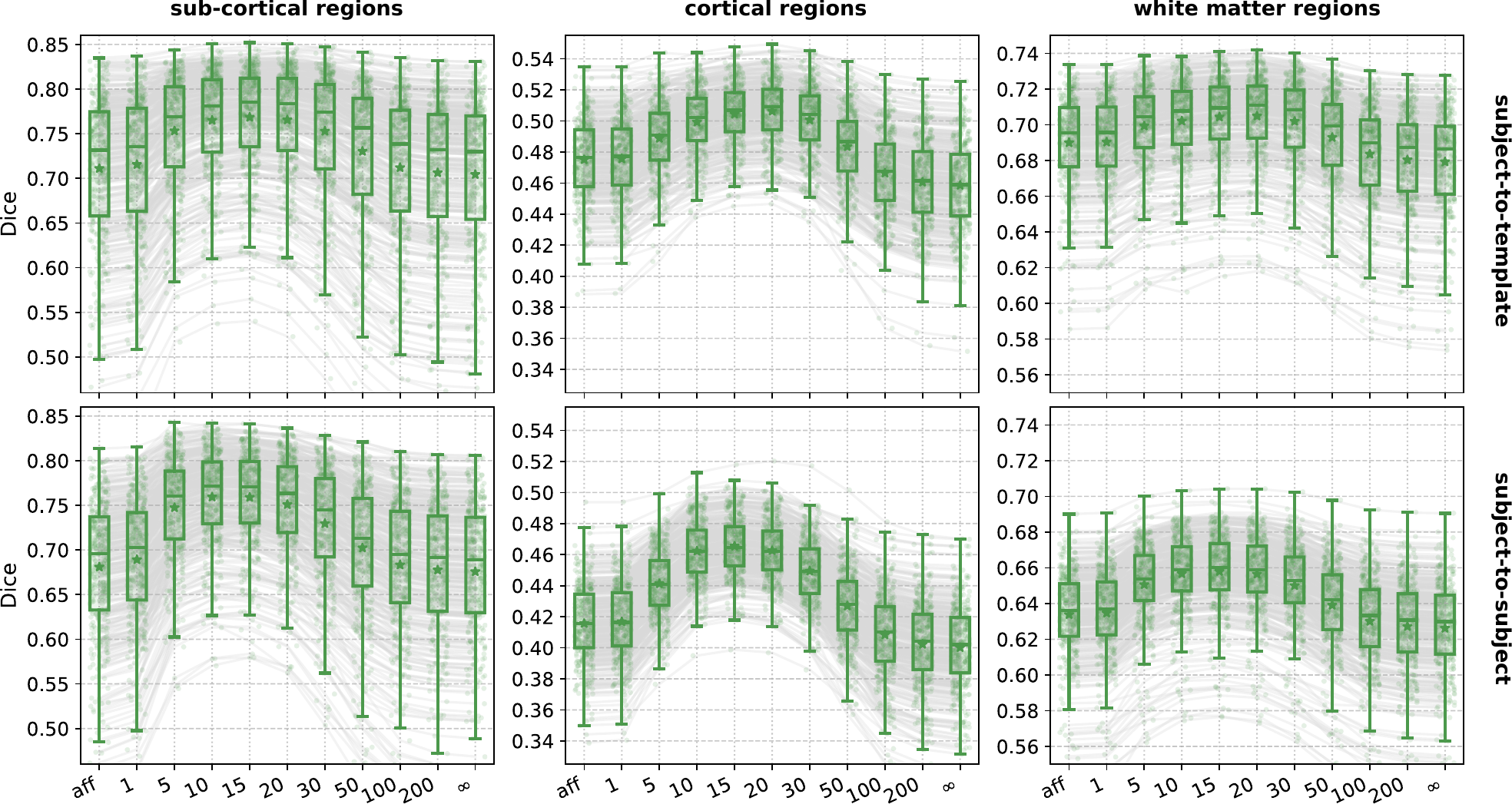}\vspace{0.1cm}
            \caption{Average Dice scores after registration with \emph{Polaffini} with various values of the smoothness parameter $\sigma$, for subject-to-template (top) and subject-to-subject (bottom).}
            \label{fig_dices_sigma}
        \end{figure}

        In this experiment, we evaluate the influence of the smoothness parameter $\sigma$ of \emph{Polaffini}. To evaluate the quality of the alignment after registration with various $\sigma$, we used an average Dice score for the three anatomical groups.
        Results are shown in Fig.~\ref{fig_dices_sigma}.
        
        Optimal overlap was achieved for $\sigma=20$ mm for subject-to-template registration and $\sigma=15$ mm for subject-to-subject one.
        
        The optimal value for $\sigma$ is, however, completely dependent on the choice of the segmentation scheme. The values are here adapted for the DKT brain segmentation. For larger regions, leading to more space between the control points, a higher value for $\sigma$ would be better suited. And vice-versa for smaller regions.
        For low values of $\sigma$, the background weight is predominant as soon as one deviate a little from the neighbourhood centers, leading to results very similar to when using the background transformation only.
        
        Interestingly, although \emph{Polaffini-aff} (background affine transformation only) and \emph{Polaffini} with $\sigma=\infty$ both yield both an affine transformation, they lead to quite important Dice score differences. \emph{Polaffini-aff}, which is similar to what is done in~\cite{Iglesias2023}, leads to better alignment compared its affine counterpart using $\sigma=\infty$.
        % \begin{figure}
        %     \centering
        %     \textbf{subject-to-template}\\
        %     \includegraphics[width=0.9\linewidth]{imgs/dices_sigma.pdf}\vspace{0.1cm}
        %     \textbf{subject-to-subject}\\
        %     \includegraphics[width=0.9\linewidth]{imgs/dices_sigma-pair.pdf}
        %     \caption{Average Dice scores after registration with Polaffini with various values of the smoothness parameter $\sigma$.}
        %     \label{fig_dices_sigma}
        % \end{figure}

    \subsubsection{Robustness} \label{robust}
\begin{figure}[h!]
        \centering
        \small
        \setlength{\tabcolsep}{2pt}
        \begin{tabular}{cccccc}
             & \textbf{reference} & \textbf{moving} & \textbf{moved} & \textbf{reference and moved}\vspace{0.1cm}\\ 
            \multirow{2}{*}{a.} & {\scriptsize ukb\_1154012-20252} & {\scriptsize adni\_AD-012-S-0720} &  \multicolumn{2}{c}{\small registered with Flirt}  \\
             & \includegraphics[width=0.21\linewidth]{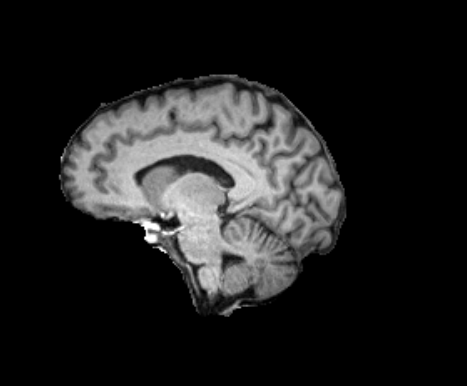} &
            \includegraphics[width=0.21\linewidth]{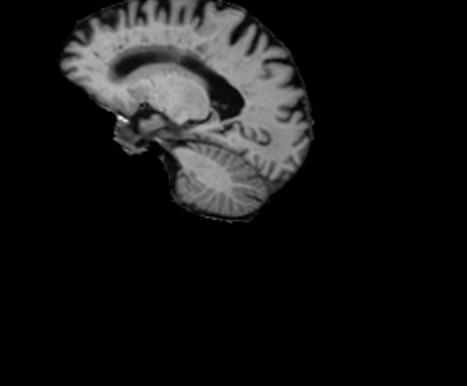} &
            \includegraphics[width=0.21\linewidth]{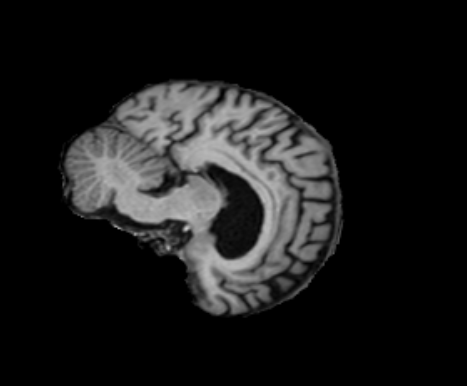} &
            \includegraphics[width=0.21\linewidth]{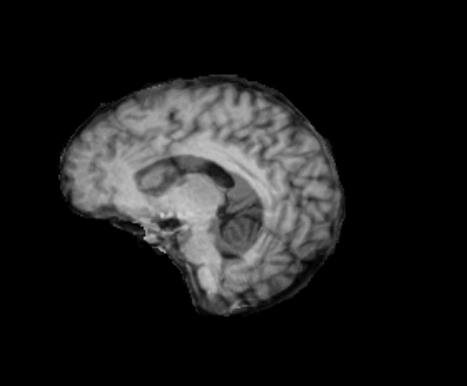} \vspace{0.1cm}\\
            \multirow{2}{*}{b.} & {\scriptsize ixi\_165-HH-1589}   & {\scriptsize ukb\_1145033-20252} & \multicolumn{2}{c}{\small registered with Flirt} \\
             & \includegraphics[width=0.21\linewidth]{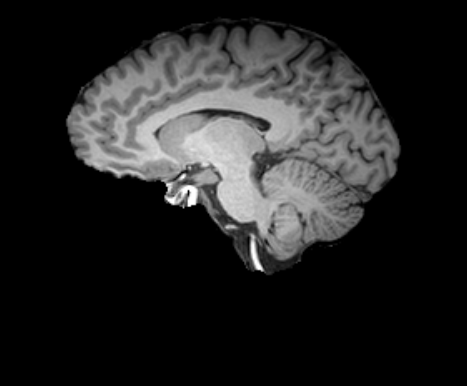} &
            \includegraphics[width=0.21\linewidth]{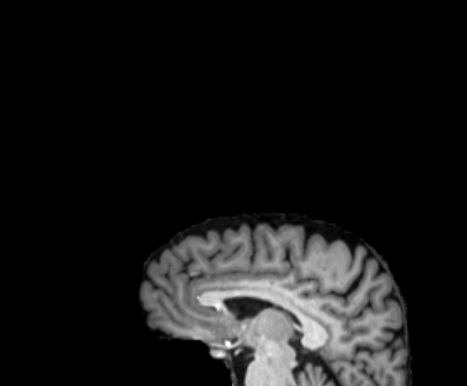} &
            \includegraphics[width=0.21\linewidth]{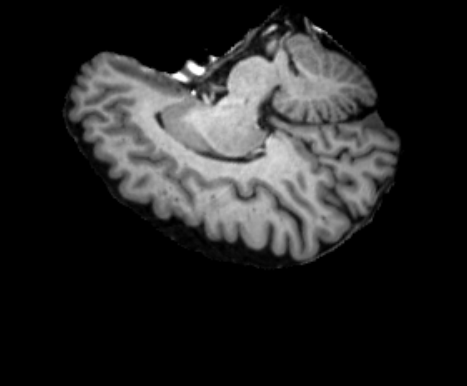} &
            \includegraphics[width=0.21\linewidth]{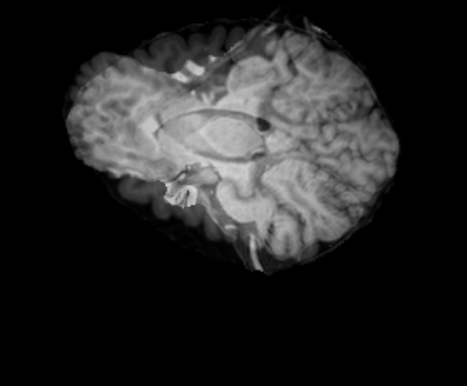} \vspace{0.1cm}\\
            \multirow{2}{*}{c.} & {\scriptsize ixi\_143-Guys-0785} & {\scriptsize adni\_MCI-003-S-6258} &  \multicolumn{2}{c}{\small registered with Anima-aff}  \\
             & \includegraphics[width=0.21\linewidth]{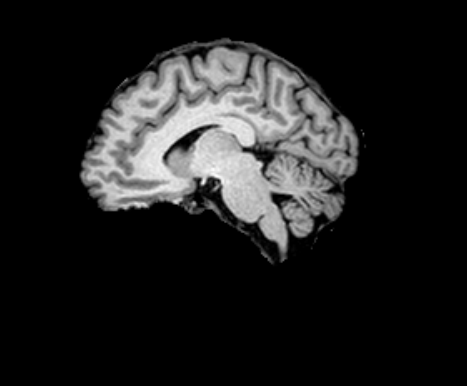} &
            \includegraphics[width=0.21\linewidth]{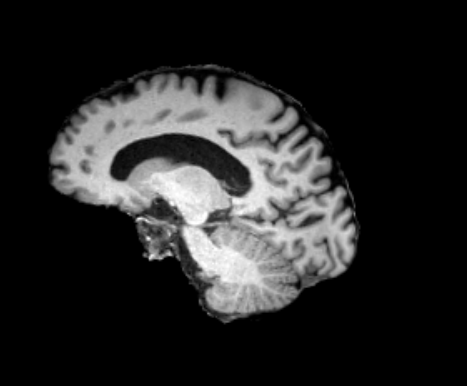} &
            \includegraphics[width=0.21\linewidth]{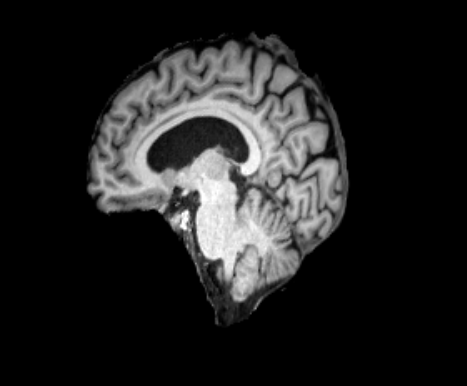} &
            \includegraphics[width=0.21\linewidth]{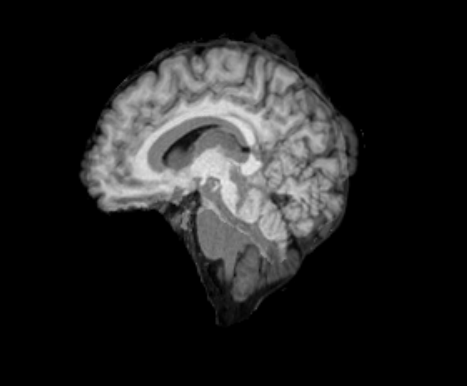}         
        \end{tabular}
        \setlength{\tabcolsep}{6pt}
        \caption{Example of failure cases. Images are shown in the grid of the reference image.}
        \label{fig_fails}
        \end{figure}

        In this experiment, we evaluate the robustness of \emph{Polaffini} and the competing methods by computing a failure rate.
        We considered as potential outliers all cases for which the average Dice score over all regions after registration was below 0.34 (Z-score below -1.5, pooling Dices from all affine methods). All potential outliers were visually inspected to assess the reason behind the poor anatomical structure overlap score. All of them were clear failure cases, not just a residual misalignment to be expected after an affine registration. The failure counts for each method are reported in table~\ref{tabfail}.
        Almost all of the failure cases only occurred when registering with Flirt, mostly when subjects from the IXI dataset were involved. The most common failure type, showcased in Fig~\ref{fig_fails}-a., consists of a local minimum where the frontal part of the moving brain is matched with the cerebellum of the reference one, a less frequent upside-down outcome is showcased in Fig~\ref{fig_fails}-b. We suspect it is due to the fact that images from IXI have their axes ordered differently compared to the other two datasets and the template. However, since this information is contained in the header, it is unclear why Flirt could not handle the situation properly. Without the restriction of the angle search to $[-30, 30]$, Flirt would give worse results (19 failure cases for subject-to-template and 61 for subject-to-subject). The only failure case using Anima for subject-to-subject is shown in Fig~\ref{fig_fails}-c. For the rest of this section, we discard registrations if at least one of the methods has failed.
        
        \begin{table}
            \centering
            \renewcommand{\arraystretch}{1.3}
            \begin{tabular}{l|cccccc|}
            \cline{2-7}
              & Flirt & ANTs-aff & Anima-aff & Aladin & Polaffini-aff & Polaffini-polyaff \\ \hline
            \multicolumn{1}{|l|}{subject-to-template} & 4.86 &  0  &  0.29  & 0  & 0 & 0 \\
            \multicolumn{1}{|l|}{subject-to-subject} & 11.71 &  0  &  0.29  & 0  & 0 & 0 \\ \hline
            \end{tabular}

            \label{tabfail}
            \caption{Failure rate (\%) for each registration method.}
        \end{table}

    \subsubsection{Anatomical structure overlap after registration} \label{regresults}
    \label{dice}
\begin{figure}
            \centering
            \includegraphics[width=\linewidth]{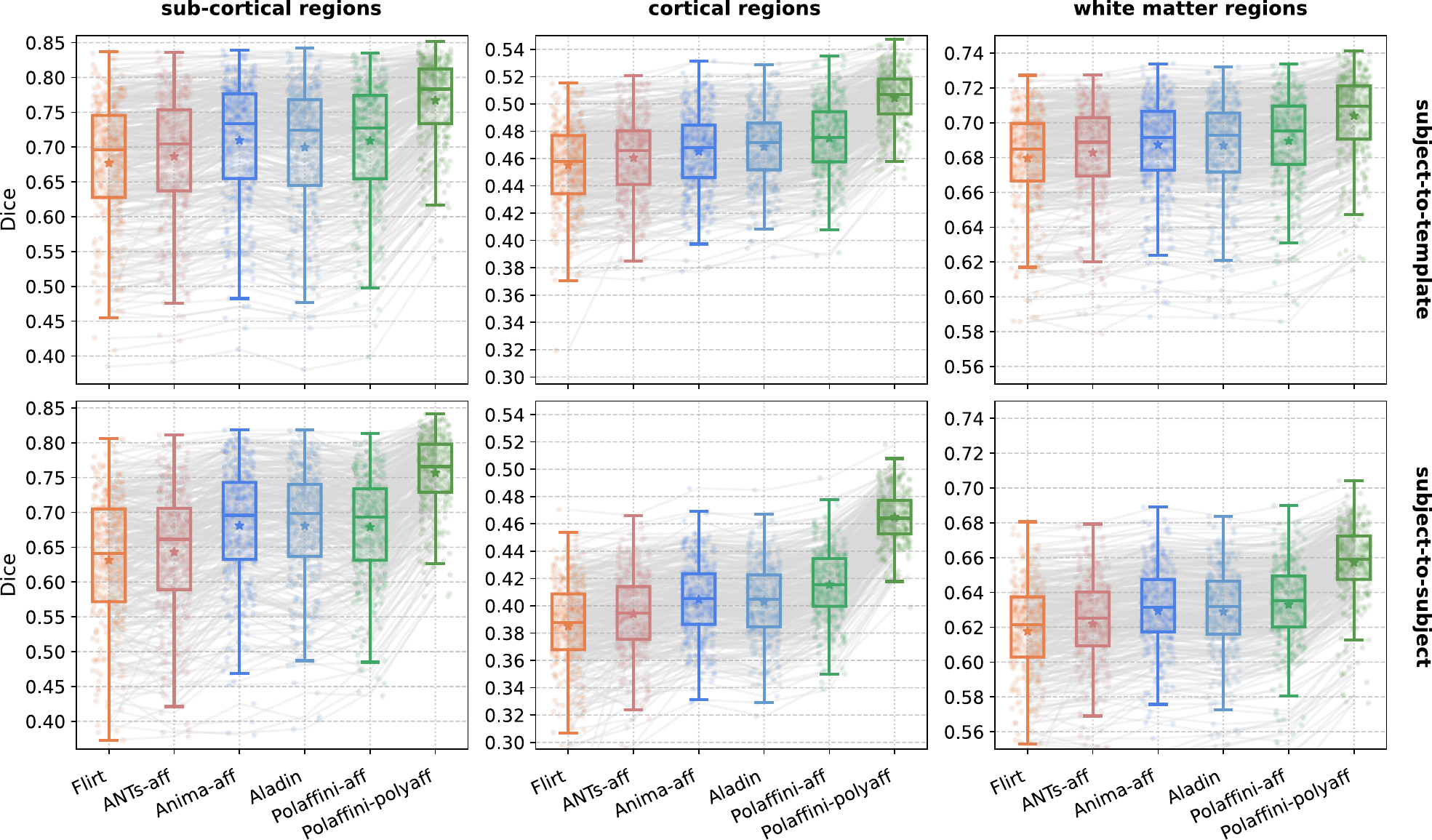}
            \caption{Average Dice scores after registration with \emph{Polaffini} and various affine registration methods, for subject-to-template (top) and subject-to-subject (bottom). colours are linked to the similarity metric types (reddish: image-wise, bluish: block-wise, greenish: \emph{Polaffini}).}
            \label{fig_dices}
        \end{figure}

        In this experiment, we evaluate the quality of the alignment after registration with \emph{Polaffini} and with the competing methods using an average Dice score for the three anatomical groups. Boxplots of those Dice scores are displayed in Fig~\ref{fig_dices}. We also recorded the frequency at which each method performs $1^\text{st}$ or $2^\text{nd}$, shown in Table~\ref{tabrankdice}. Pairwise Wilcoxon signed-rank tests ($\alpha = 0.05$, Bonferroni correction) were conducted to assess statistical significance; raw p-values are reported in Table~\ref{tabstatdice}, with significant results highlighted in colour.
        
        The polyaffine version of \emph{Polaffini} leads to much better structural alignment in all anatomical groups compared to the other methods that estimate an affine transformation. Among those affine registration methods, \emph{Polaffini-aff} globally achieves better performances, although Anima-aff and Aladin are relatively close for sub-cortical structures. 
        % \begin{figure}
        %     \centering
        %     \textbf{subject-to-template}\\
        %     \includegraphics[width=\linewidth]{imgs/dices.pdf}\vspace{0.1cm}
        %     \textbf{subject-to-subject}\\
        %     \includegraphics[width=\linewidth]{imgs/dices-pair.pdf}
        %     \caption{Average Dice scores after registration with Polaffini and various affine registration methods.}
        %     \label{fig_dices}
        % \end{figure}
        
        \begin{table}
            \centering
            \renewcommand{\arraystretch}{1.2}
            \begin{tabular}{l|l|l|cccccc|}
                \cline{2-9}
                & regions & rank & Flirt & ANTs-aff & Anima-aff & Aladin & Polaffini-aff & Polaffini-polyaff \\ \hline
               \multicolumn{1}{|c|}{\multirow{3}{*}{\rotatebox{90}{{subject-to-template\ }}}} & \multirow{2}{*}{sub-cortical}  & 1$^{\mathrm{st}}$ &  0.3 &  0.  &  0.9  & 1.2  & 0.6 & \textbf{96.99} \\
                \multicolumn{1}{|l|}{}& & 2$^{\mathrm{nd}}$ &  1.51 & 3.61 & 34.04 & 15.66 & \textbf{43.37} & 1.81 \\ \cdashline{2-9}
                \multicolumn{1}{|l|}{}& \multirow{2}{*}{cortical} & 1$^{\mathrm{st}}$ & 0 & 0 & 0. & 0. & 0 & \textbf{100}\\
                \multicolumn{1}{|l|}{}& & 2$^{\mathrm{nd}}$ & 2.11 & 1.2 & 6.02 & 12.35 & \textbf{78.31} & 0 \\ \cdashline{2-9}
                \multicolumn{1}{|l|}{}& \multirow{2}{*}{white matter} & 1$^{\mathrm{st}}$ & 0 & 0 & 0 & 0.3 & 2.11 & \textbf{97.59}\\
                \multicolumn{1}{|l|}{}& & 2$^{\mathrm{nd}}$ & 0.9 & 1.81 & 22.59 & 16.27 & \textbf{56.93} & 1.51 \\ \hline
                
                \multicolumn{1}{|l|}{\multirow{3}{*}{\rotatebox{90}{subject-to-subject\quad}}} & \multirow{2}{*}{sub-cortical}  & 1$^{\mathrm{st}}$ & 0  &  0 &   0.97 & 0.65 & 0  & \textbf{98.38}\\
                \multicolumn{1}{|l|}{}& & 2$^{\mathrm{nd}}$ &  1.95 & 2.92 & 32.14 & 28.9 & \textbf{33.12} & 0.97 \\ \cdashline{2-9}
                \multicolumn{1}{|l|}{}& \multirow{2}{*}{cortical} & 1$^{\mathrm{st}}$ & 0 & 0 & 0 & 0 & 0 & \textbf{100}\\
                \multicolumn{1}{|l|}{}& & 2$^{\mathrm{nd}}$ &  0 &  0.97& 7.79 & 3.9 & \textbf{87.34} & 0 \\ \cdashline{2-9}
                \multicolumn{1}{|l|}{}& \multirow{2}{*}{white matter} & 1$^{\mathrm{st}}$ & 0 & 0 & 0 &   0 & 0& \textbf{100}\\
                \multicolumn{1}{|l|}{}& & 2$^{\mathrm{nd}}$ &1.62 & 0 & 16.56 &12.34& \textbf{69.48} & 0 \\ \hline
            \end{tabular}
            \renewcommand{\arraystretch}{1}
            \label{tabrankdice}
            \caption{Frequency (\%) at which each method performs 1$^{\mathrm{st}}$ and 2$^{\mathrm{nd}}$ in terms of Dice score, for each registration, averaged by label.}
        \end{table}
        \begin{table}
            \centering
            \begin{tabular}{|ll||rrr|rrr|}
            \hline
            \multicolumn{2}{|c||}{\textbf{methods}} & \multicolumn{3}{c|}{\textbf{subject-to-template}} & \multicolumn{3}{c|}{\textbf{subject-to-subject}} \\
            A & B & sub-cort & cort & WM & sub-cort & cort & WM \\ \hline
            ANTs-aff & Flirt & {\color{blue}1.0e-23} & {\color{blue}1.1e-18} & {\color{blue}4.0e-20} & {\color{blue}3.5e-10} & {\color{blue}6.0e-22} & {\color{blue}5.5e-17}\\
            Anima-aff & Flirt & {\color{blue}6.5e-52} & {\color{blue}1.8e-31} & {\color{blue}1.1e-48} & {\color{blue}1.0e-42} & {\color{blue}3.8e-42} & {\color{blue}4.9e-44}\\
            Anima-aff & ANTs-aff & {\color{blue}4.5e-46} & {\color{blue}6.0e-17} & {\color{blue}3.1e-36} & {\color{blue}6.6e-43} & {\color{blue}2.8e-33} & {\color{blue}6.8e-40}\\
            Aladin & Flirt & {\color{blue}8.3e-35} & {\color{blue}6.8e-49} & {\color{blue}1.9e-46} & {\color{blue}8.4e-43} & {\color{blue}2.1e-40} & {\color{blue}1.1e-46}\\
            Aladin & ANTs-aff & {\color{blue}1.3e-19} & {\color{blue}3.1e-45} & {\color{blue}4.4e-38} & {\color{blue}2.5e-41} & {\color{blue}2.6e-28} & {\color{blue}4.7e-44}\\
            Aladin & Anima-aff & {\color{red}1.7e-19} & {\color{blue}8.1e-17} & {\color{gray}7.9e-02} & {\color{gray}3.9e-01} & {\color{red}5.0e-06} & {\color{gray}1.5e-03}\\
            Polaffini-aff & Flirt & {\color{blue}1.3e-49} & {\color{blue}4.0e-55} & {\color{blue}2.8e-53} & {\color{blue}7.6e-39} & {\color{blue}3.6e-52} & {\color{blue}3.6e-51}\\
            Polaffini-aff & ANTs-aff & {\color{blue}5.6e-39} & {\color{blue}1.2e-54} & {\color{blue}1.8e-47} & {\color{blue}5.9e-31} & {\color{blue}5.3e-52} & {\color{blue}7.6e-51}\\
            Polaffini-aff & Anima-aff & {\color{gray}9.2e-01} & {\color{blue}4.9e-48} & {\color{blue}7.2e-18} & {\color{gray}5.2e-02} & {\color{blue}3.2e-47} & {\color{blue}6.2e-25}\\
            Polaffini-aff & Aladin & {\color{blue}8.1e-13} & {\color{blue}6.1e-40} & {\color{blue}1.6e-22} & {\color{gray}3.2e-01} & {\color{blue}3.1e-50} & {\color{blue}9.1e-31}\\
            Polaffini-polyaff & Flirt & {\color{blue}3.7e-56} & {\color{blue}3.6e-56} & {\color{blue}3.6e-56} & {\color{blue}3.0e-52} & {\color{blue}3.0e-52} & {\color{blue}3.0e-52}\\
            Polaffini-polyaff & ANTs-aff & {\color{blue}3.9e-56} & {\color{blue}3.6e-56} & {\color{blue}3.6e-56} & {\color{blue}3.0e-52} & {\color{blue}3.0e-52} & {\color{blue}3.0e-52}\\
            Polaffini-polyaff & Anima-aff & {\color{blue}5.1e-56} & {\color{blue}3.6e-56} & {\color{blue}3.8e-56} & {\color{blue}5.0e-52} & {\color{blue}3.0e-52} & {\color{blue}3.0e-52}\\
            Polaffini-polyaff & Aladin & {\color{blue}5.1e-56} & {\color{blue}3.6e-56} & {\color{blue}3.9e-56} & {\color{blue}7.5e-52} & {\color{blue}3.0e-52} & {\color{blue}3.0e-52}\\
            Polaffini-polyaff & Polaffini-aff & {\color{blue}4.7e-56} & {\color{blue}3.6e-56} & {\color{blue}5.6e-56} & {\color{blue}3.0e-52} & {\color{blue}3.0e-52} & {\color{blue}3.0e-52}\\\hline
            \end{tabular}
            \label{tabstatdice}
            \caption{Pairwise statistical comparison of average Dice scores after image registration using \emph{Polaffini} and various affine registration methods. Raw p-values for Wilcoxon signed-rank tests ($\alpha=0.05$). Significant comparisons after Bonferroni correction in favor of method A (resp. method B) are highlighted in blue (resp. red); non-significant ones in gray.}
        \end{table}
                
    \subsubsection{Geometric comparison of the estimated affine transformations} \label{geocomp}

    In this experiment, we analyse the geometric differences between the affine transformations estimated by \emph{Polaffini-aff} and those produced by conventional affine registration methods. Since this comparison focuses on affine matrices, the polyaffine variant of \emph{Polaffini} is excluded.
    As established in Section~\ref{dice}, \emph{Polaffini-aff} generally achieves better alignment of anatomical structures. Accordingly, we use it as the reference and compute distances from each alternative method to \emph{Polaffini-aff}, decomposing the affine transformations into interpretable geometric components.

\begin{figure}
        \centering
        \includegraphics[width=\linewidth]{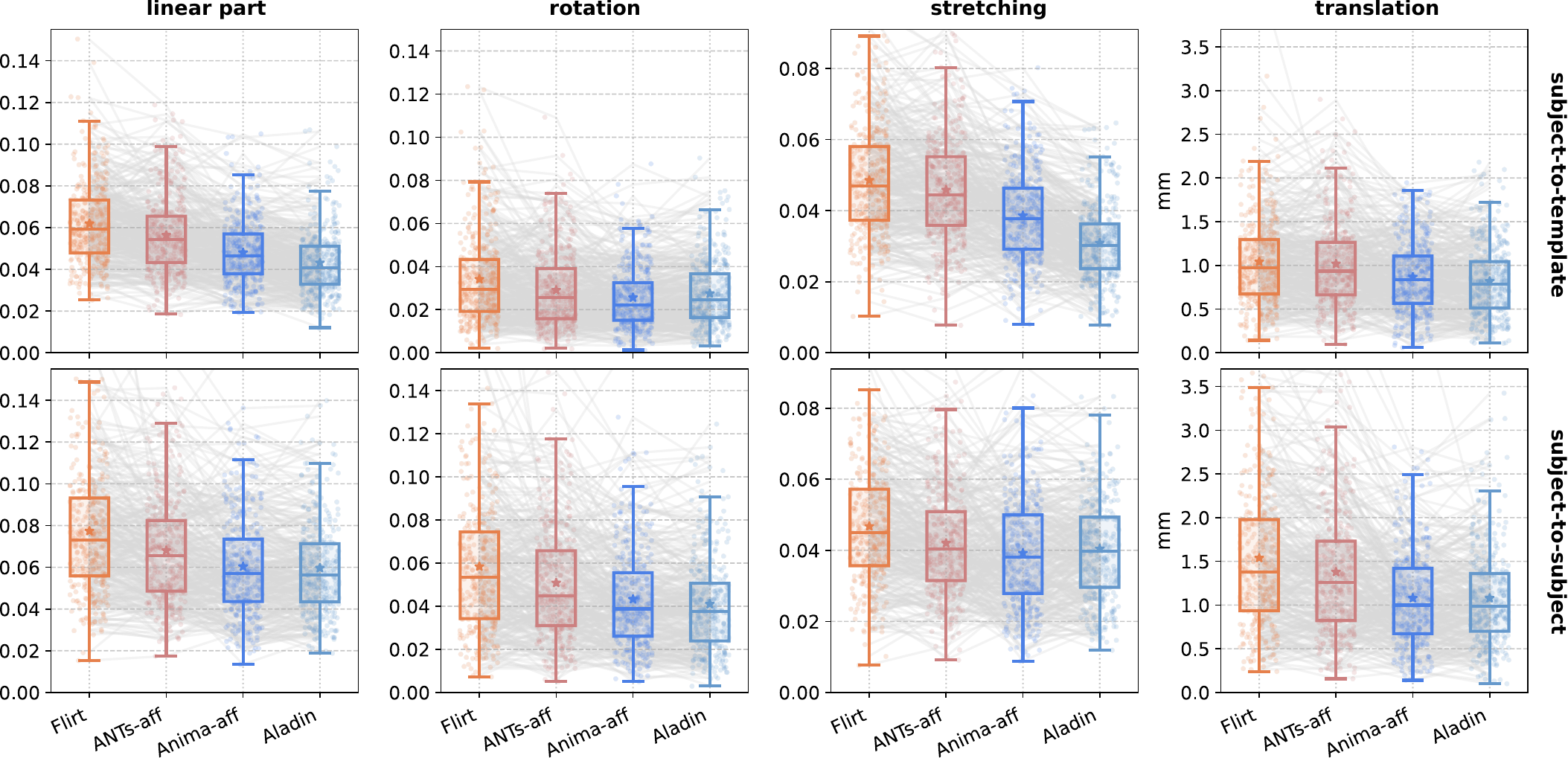}
        \caption{Geometric differences between the affine transformations estimated by \emph{Polaffini-aff} and the ones from various affine registration methods, for subject-to-template (top) and subject-to-subject (bottom).}
        \label{fig_affdists}
    \end{figure}

    An affine transformation on Euclidean $d$-space acts on a point $x \in \mathbb{R}^d$ as $y = Lx + t$, where $L \in \operatorname{GL}(d, \mathbb{R})$ is the linear part and $t \in \mathbb{R}^d$ is the translation vector. By the polar decomposition, the linear part can be written as $L = RS$, where $R \in \operatorname{SO}(d)$ is a rotation matrix and $S \in \operatorname{Sym}^+(d, \mathbb{R})$ is a symmetric positive-definite matrix capturing the stretching component. Given two affine transformations, indexed by $i$ and $j$, we compute distances between their components using metrics adapted to the geometric nature of each space, as summarised in Table~\ref{matdists}.
    \begin{table}
    \centering
    \renewcommand{\arraystretch}{1.5}
    \begin{tabular}{|l|c|ccc|}
    \hline
    component & set & \multicolumn{3}{c|}{geometric distance}\\ \hline
    linear part & $\operatorname{GL}(d,\mathbb{R})$ & $\mathrm{dist}(L_i, L_j)$ &=& $\left\|\log\left(L_i^{-1}L_j\right)\right\|_F$ \\
    rotation & $\operatorname{SO}(d)$ & $\mathrm{dist}(R_i, R_j)$ &=& $\left\|\log\left(R_i^TR_j\right)\right\|_F$ \\
    stretch & $\operatorname{Sym}^+(d, \mathbb{R})$ & $\mathrm{dist}(S_i, S_j)$ &=&  $\left\|\log\left(S_i^{-1/2}S_j S_i^{-1/2}\right)\right\|_F$ \\
    translation & $\mathbb{R}^d$ & $\mathrm{dist}(t_i, t_j)$ &=&  $\left\|t_i-t_j\right\|_2$\\ \hline
    \end{tabular}
    \renewcommand{\arraystretch}{1}
    \caption{Geometric distances for the affine matrix components. $\log$ is the matrix logarithm and $\|.\|_F$ is the Frobenius norm.}
    \label{matdists}
    \end{table}
    Given affine transformations estimated by two of the evaluated affine registration methods, we computed the above-mentioned geometry-aware distances between their components. Boxplots of those distances are shown in Fig.~\ref{fig_affdists}.
    
    We observe that block-matching approaches (Anima-aff and Aladin) produce affine transformations that are usually geometrically closer to \emph{Polaffini-aff} compared to their counterparts using image-wise similarity metrics (Flirt and ANTs-aff).
    % \begin{figure}
    %     \centering
    %     \textbf{subject-to-template}\\
    %     \includegraphics[width=\linewidth]{imgs/affdists.pdf}\vspace{0.3cm}
        
    %     \textbf{subject-to-subject}\\
    %     \includegraphics[width=\linewidth]{imgs/affdists-pair.pdf}
    %     \caption{Geometric differences between the affine transformations estimated by Polaffini-aff and the ones from various affine registration methods.}
    %     \label{fig_affdists}
    % \end{figure}

    \subsubsection{Polaffini as pre-alignment for non-linear registration} \label{nonlininit}
    
        In this section, we evaluate the ability of \emph{Polaffini} to improve the results of subsequent traditional or deep-learning non-linear registration methods depicted in Section~\ref{nlintools}. The evaluation was performed on the subjects from the testing set that were unseen by the deep-learning models at training. To evaluate the quality of the alignment after the non-linear registration, we used an average Dice score for the three anatomical groups. In order to be comparable with~\cite{Dalca2019}, we also reported results for cortex as a whole, where the cortical regions have been fused into one label and single Dice was computed.
        \begin{remark}
            The results presented in Section~\ref{nonlininit} were produced for the anterior publication associated with this work~\citep{Legouhy2023}. These results were obtained using the original parameter settings described in~\citep{Legouhy2023}; we did not re-run those experiments with the new parameters. In particular, \emph{Polaffini} used a smoothness parameter of $\sigma=20$ instead of $15$, and segmentations were obtained using FastSurfer instead of SynthSeg.
        \end{remark}
       
        \paragraph{Intensity-based affine against Polaffini-polyaff pre-alignment}\mbox{}\\
        
\begin{figure}
            \centering
            \includegraphics[width=\linewidth]{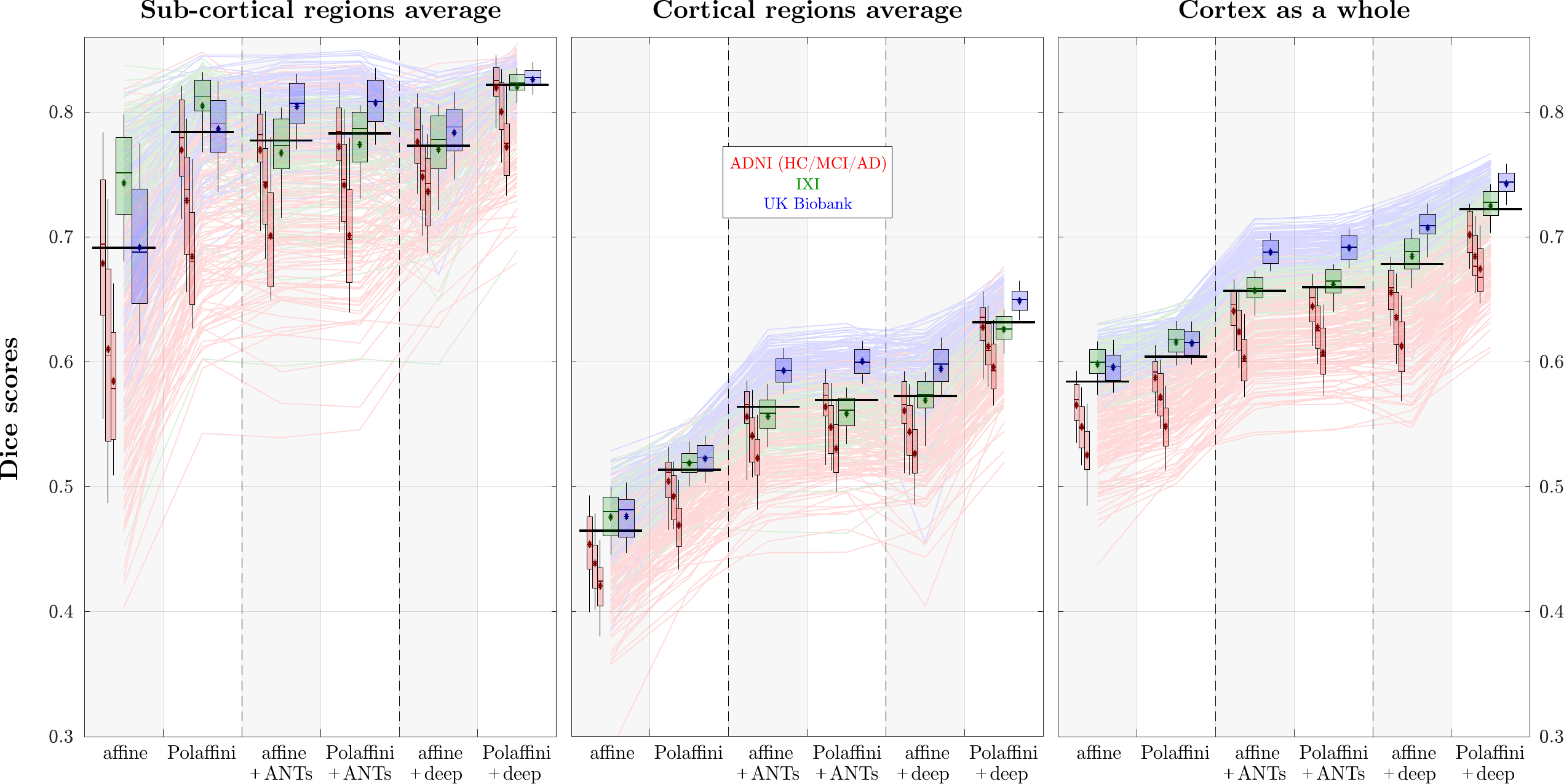}
            \caption{Dice scores after non-linear registration with traditional (ANTs) and deep-learning approaches, initialised with intensity-based affine (affine) and \emph{Polaffini-polyaff} (Polaffini).}
            \label{dicefig}
        \end{figure}

        In this experiment we compare the results after non-linear registration when initialising with an intensity-based affine registration from Flirt against initialising with \emph{Polaffini-polyaff}. Boxplots of the Dice scores are shown in Fig.~\ref{dicefig}. Statistics regarding differences between overlap scores after registration with affine against proposed polyaffine initialisation are reported in Table~\ref{tabstat}. 
        \begin{table}
            \renewcommand{\arraystretch}{1.1}
            \setlength{\tabcolsep}{5pt}
            \centering
            \begin{tabular}{c||l|l|l|c}
            & affine init. & polyaffine init. & p-value & Cohen's d\\\hline\hline
            \multirow{3}{*}{\begin{tabular}{c}sub-cortical\\ regions\end{tabular}} & affine &  polyaffine &  $< 10^{-115}$ & 1.864 \\
            & affine + ANTs &  polyaffine + ANTs &  $< 10^{-14}$ & 0.447  \\
            & affine + deep &  polyaffine + deep &  $< 10^{-108}$ & 1.762 \\\hline
            \multirow{3}{*}{cortex} & affine &  polyaffine & $< 10^{-107}$ & 1.745 \\
            & affine + ANTs &  polyaffine + ANTs & $< 10^{-38}$ & 0.798 \\
            & affine + deep &  polyaffine + deep & $< 10^{-176}$ &  3.019 \\
            \end{tabular}
            \caption{Statistics about differences between overlap scores after registration with affine vs proposed polyaffine initialisation. Reported p-values are for paired t-tests, Cohen's d for paired samples (with polyaffine $-$ affine on numerator).}
            \label{tabstat}
        \end{table}
        
\begin{figure}
            \centering
            \includegraphics[width=0.85\linewidth]{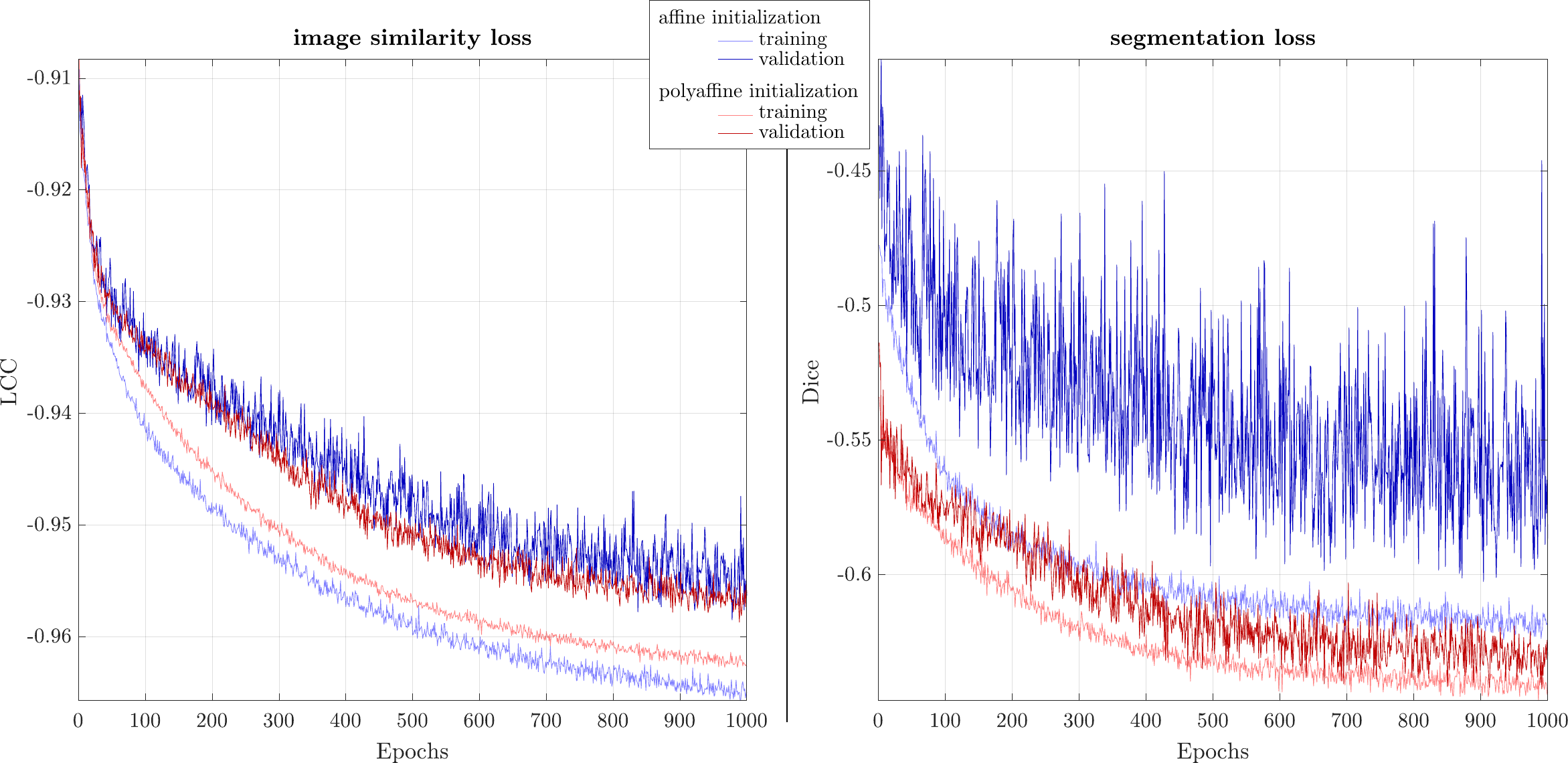}
            \caption{Evolution of image similarity (LCC) and segmentation (average Dice on all regions) losses during training of deep-learning models with affine (blue) and proposed polyaffine (red) initialisation for training (light) and validation (dark) samples.}
            \label{lossfig}
        \end{figure}

        For traditional non-linear registration, we only observe a limited improvement of the alignment when initialising with the proposed polyaffine method. For the deep-learning non-linear registration however, we observe much better alignment when using \emph{Polaffini-polyaff} as pre-alignment.
        Part of the explanation for why the proposed polyaffine initialisation leads to better results in the deep-learning case can be found by examining the evolution of the losses during training (see Fig.~\ref{lossfig}). While image similarity losses follow similar trajectories for both approaches, segmentation losses, which actually quantifies the alignment of anatomical structures, show a much smoother profile with the proposed polyaffine starting point. The gap between the training and validation losses is also much smaller with the proposed initialisation.

        \paragraph{Influence of the initialisation for training and testing}\mbox{}\\

\begin{figure}
            \centering
            \includegraphics[width=\linewidth]{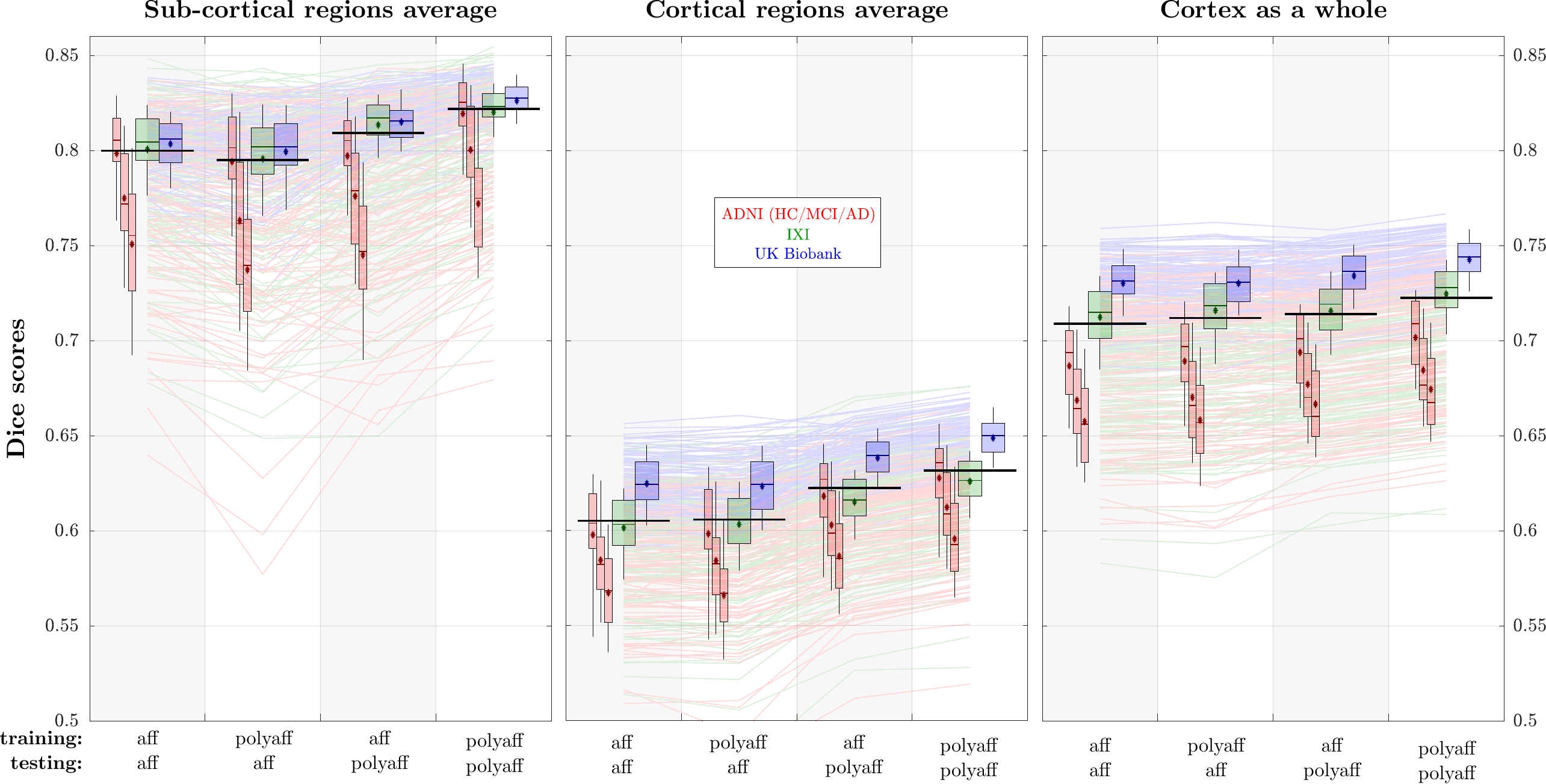}
            \caption{Dice scores after deep-learning non-linear registration for different configurations of initialisation with \emph{Polaffini-aff} and \emph{Polaffini-polyaff} at training and testing.}
            \label{compofig}
        \end{figure}

        In this experiment we compare the results after deep-learning non-linear registration when initialising with various combinations of \emph{Polaffini-aff} and \emph{Polaffini-polyaff} during training and testing. This choice is motivated by the emergence of the paper from~\cite{Iglesias2023} in which an initialisation similar to \emph{Polaffini-aff} is used both for training and testing. Boxplots of the Dice scores are shown in Fig.~\ref{compofig}.      
        The best results are obtained when using \emph{Polaffini-polyaff} as initialisation both for training and testing. The second best results are obtained when using \emph{Polaffini-aff} at training but \emph{Polaffini-polyaff} at testing. In this case, the spatial distribution at testing is contained in the one seen at training. This is not the case when using \emph{Polaffini-polyaff} at training but affine at testing, which leads to poorest results for the sub-cortical regions compared to when using affine for both.
    
\section{Discussion}

    To help avoid local minima at the finest resolution, most traditional non-linear registration algorithms follow a coarse-to-fine resolution optimisation and some recent deep-learning models also adopt such pyramidal strategies at training~\citep{mok2020}. However, they still rely on a prior, usually intensity-based, affine alignment to recover the largest displacement. Although it may allow a user to skip the first coarse levels of the pyramid, our method should not be seen as a rival to the coarse-to-fine approach but as a more robust, anatomically grounded and less constrained alternative to the affine pre-alignment. 
    
    In~\cite{Commowick2008}, an anatomically grounded non-linear registration scheme was proposed where optimal affine transformations between homologous delineated regions were sought by iteratively optimising an image similarity criterion. An overall transformation is then constructed by attributing the estimated affine displacement to an eroded version of the associated regions and using a log-Euclidean interpolation for a smooth transition to enforce a diffeomorphic result. Our approach, by contrast, does not require slow iterative optimisation for the matching and is intended to be taken as an alternative to affine initialisation rather than a non-linear registration endpoint.

    In KeyMorph~\citep{keymorph}, a deep-learning model is trained to extract feature points, guided by an affine registration task. The model learns to predict points that maximise alignment between the fixed and moving images when an affine transformation is computed from the corresponding feature points. This framework is designed to be robust to extreme misalignment, such as very large rotations. However, these scenarios are simulated and it is unclear how they could occur in real clinical settings due to the physical constraints imposed by the scanner on subject positioning. A subsequent version~\citep{keymorph2}, enables the estimation of more flexible transformations. However, the chosen transformation model, based on thin plate splines, does not guarantee diffeomorphic properties. Additionally, the learned feature points offer limited utility for downstream analysis and lack clear anatomical interpretability. In contrast, segmentation-based feature extraction provides anatomically grounded correspondences and integrates naturally with existing neuroimaging pipelines where segmentation is already standard practice.

    To estimate local transformations beyond translations, such as local affine models, we organised the feature points into neighbourhoods based on spatial proximity, and estimated transformations between corresponding neighbourhoods. This strategy is simple and incurs minimal computational cost, since neighbourhoods can be defined once and the resulting local transformation estimates admit closed-form solutions. Yet, to the best of our knowledge, existing keypoint-based and block-matching registration algorithms do not explicitly leverage such neighbourhood-based estimation; we believe it could be integrated into them straightforwardly to enable richer transformation models with negligible overhead.

    Assuming that the segmentation process was successful, quality control of the registration step can easily be done just by computing an overlap measure between the transformed moving and reference segmentations. It has been shown to be an efficient way of detecting failure cases in Section~\ref{robust}.

    Since \emph{Polaffini} relies on feature points extracted from segmentations, it inherits the generalisation capabilities of the segmentation tool employed. In particular, when using SynthSeg~\citep{Billot2023}, which is trained with highly aggressive data augmentation, the resulting feature extraction becomes agnostic to image contrast and resolution.
    
    We used segmentations based on the DKT atlas that SynthSeg~\citep{Billot2023} and FastSurfer~\citep{Henschel2020} pre-trained models are designed to output. This may, however, be sub-optimal as some regions are quite large (e.g. single label for superior frontal) leading to a low density of feature points in some areas. The method would likely benefit from more fine-grained and equi-distributed segmentations such as gyri delineation.

\section{Conclusion}
    In this paper, we presented \emph{Polaffini}, a versatile registration framework that leverages anatomical segmentations, now readily available via pre-trained deep-learning models. \emph{Polaffini} enables the estimation of both affine and polyaffine transformations that are anatomically grounded. Polyaffine transformations offer significantly more degrees of freedom than their affine counterparts, allowing for finer alignment while remaining embedded in a mathematically well-defined framework. The method relies on global and local affine estimations with closed-form solutions, so it is fast. We demonstrated that \emph{Polaffini} is highly robust and achieves improved anatomical alignment compared to widely used intensity-based alternatives. Furthermore, it provides a better initialisation for non-linear registration, leading to enhanced alignment at the end-point, particularly in a deep-learning setting.

% \subsection{Front Matter}

% Please use the below tags for the article front matter:

\section*{Data and Code Availability}

The code for \emph{Polaffini} is available open-source, with documentation and examples, at~\url{https://github.com/CIG-UCL/polaffini}.

No new data has been acquired for this paper. The data used comes from various datasets listed in~\ref{data}, with more information provided in the Acknowledgements section. %~\ref{acks}.

\section*{Author Contributions}

A.L.: Conceptualisation, Methodology, Validation, Software (lead dev.), Writing (original draft).\\
C.C.: Validation, Writing (review), Software (contributor).\\
R.C.: Methodology (through discussion), Software (contributor).\\
H.A.: Resources.\\
H.Z.: Supervision, Conceptualisation, Writing (review and editing).

\section*{Funding}

H.A. was supported by UKRI Future Leaders Fellowship MR/W011980/1 -- QUANTIMA: Quantitative imaging platform for the diagnosis, subtyping, staging and outcome prognosis in dementia.\\
A.L., R.C., H.A., and H.Z. were supported by Innovate UK grant 10036158 -- CLAIR: first-in-class non-invasive test for outcome prediction and stratification in dementia.\\
C.C. was supported by the EPSRC-funded EP/S021930/1: UCL Centre for Doctoral Training in Intelligent, Integrated Imaging in Healthcare (i4health), and Brain Research UK 157806.

\section*{Declaration of Competing Interests}

We declare we do not have conflicts of interest.

\section*{Acknowledgements}
\label{acks}
About the datasets used in this paper:
\begin{itemize}
    \item  For more information about the IXI dataset, go to: \url{https://brain-development.org/ixi-dataset/}.

    \item UK Biobank is a large-scale biomedical database and research resource containing genetic, lifestyle and health information from half a million UK participants. UK Biobank’s database, which includes blood samples, heart and brain scans and genetic data of the 500,000 volunteer participants, is globally accessible to approved researchers who are undertaking health-related research that’s in the public interest. UK Biobank recruited 500,000 people aged between 40-69 years in 2006-2010 from across the UK. With their consent, they provided detailed information about their lifestyle, physical measures and had blood, urine and saliva samples collected and stored for future analysis. UK Biobank’s research resource is a major contributor in the advancement of modern medicine and treatment, enabling better understanding of the prevention, diagnosis and treatment of a wide range of serious and life-threatening illnesses – including cancer, heart diseases and stroke. Since the UK Biobank resource was opened for research use in April 2012, over 20,000 researchers from 90+ countries have been approved to use it and more than 2,000 peer-reviewed papers that used the resource have now been published.\\
    UK Biobank is generously supported by its founding funders the Wellcome Trust and UK Medical Research Council, as well as the British Heart Foundation, Cancer Research UK, Department of Health, Northwest Regional Development Agency and Scottish Government. The organisation has over 150 dedicated members of staff, based in multiple locations across the UK.\\
    You can find out more about UK Biobank at \url{http://www.ukbiobank.ac.uk}.

    \item Data used in preparation of this article were obtained from the Alzheimer’s Disease Neuroimaging Initiative (ADNI) database (adni.loni.usc.edu). As such, the investigators within the ADNI contributed to the design and implementation of ADNI and/or provided data but did not participate in analysis or writing of this report.
    A complete listing of ADNI investigators can be found at: \url{http://adni.loni.usc.edu/wp-content/uploads/how_to_apply/ADNI_Acknowledgement_List.pdf}\\
    Data collection and sharing for this project was funded by the Alzheimer's Disease Neuroimaging Initiative(ADNI) (National Institutes of Health Grant U01 AG024904) and DOD ADNI (Department of Defense award number W81XWH-12-2-0012). ADNI is funded by the National Institute on Aging, the National Institute of Biomedical Imaging and Bioengineering, and through generous contributions from the following: AbbVie, Alzheimer’s Association; Alzheimer’s Drug Discovery Foundation; Araclon Biotech; BioClinica, Inc.; Biogen; Bristol-Myers Squibb Company; CereSpir, Inc.; Cogstate; Eisai Inc.; Elan Pharmaceuticals, Inc.; Eli Lilly and Company; EuroImmun; F. Hoffmann-La Roche Ltd and its affiliated company Genentech, Inc.; Fujirebio; GE Healthcare; IXICO Ltd.; Janssen Alzheimer Immunotherapy Research \& Development, LLC.; Johnson \& Johnson Pharmaceutical Research \& Development LLC.; Lumosity; Lundbeck; Merck \& Co., Inc.; Meso Scale Diagnostics, LLC.; NeuroRx Research; Neurotrack Technologies; Novartis Pharmaceuticals Corporation; Pfizer Inc.; Piramal Imaging; Servier; Takeda Pharmaceutical Company; and Transition Therapeutics. The Canadian Institutes of Health Research is providing funds to support ADNI clinical sites in Canada. Private sector contributions are facilitated by the Foundation for the National Institutes of Health (\url{www.fnih.org}). The grantee organization is the Northern California Institute for Research and Education, and the study is coordinated by the Alzheimer’s Therapeutic Research Institute at the University of Southern California. ADNI data are disseminated by the Laboratory for Neuro Imaging at the University of Southern California.
\end{itemize}

% \section*{Supplementary Material}

% Supplementary Material (created during production as a web link to online material).

\printbibliography

% \appendix

% \section{Appendix}

% Appendices (optional).

\end{document}